\documentclass[10pt,twocolumn,letterpaper]{article}
\usepackage{xpatch}
\xpretocmd{\part}{\setcounter{section}{0}}{}{}
\usepackage{cvpr}
\usepackage{times}
\usepackage{textcomp}
\usepackage{epsfig}
\usepackage{graphicx}
\usepackage{amsmath}
\usepackage{amssymb}
\usepackage{hhline}
\usepackage{enumitem}
\usepackage{multirow}
\usepackage[normalem]{ulem}
\newenvironment{packed_enum}{
\begin{itemize}[leftmargin=*]
  \setlength{\itemsep}{0.6pt}
  \setlength{\parskip}{0pt}
  \setlength{\parsep}{0pt}
}{\end{itemize}}
\usepackage{lipsum} 
\usepackage{tabularx,colortbl}
\usepackage{float}
\usepackage{subcaption}
\usepackage[symbol]{footmisc}
\usepackage[figurename=Fig.]{caption}
\usepackage{makecell}
\floatstyle{plaintop}
\restylefloat{table}

\usepackage{soul} 
\newcolumntype{?}{!{\vrule width 1pt}}
\newcolumntype{P}[1]{>{\centering\arraybackslash}p{#1}}
\newcolumntype{C}[1]{>{\centering\arraybackslash}c{#1}}
\usepackage[pagebackref=true,breaklinks=true,letterpaper=true,colorlinks,bookmarks=false]{hyperref}

\newcommand\blfootnote[1]{%
  \begingroup
  \renewcommand\thefootnote{}\footnote{#1}%
  \addtocounter{footnote}{-1}%
  \endgroup
}

\cvprfinalcopy 

\def\cvprPaperID{6584} 

\ifcvprfinal\fi
\renewcommand{\thefootnote}{\fnsymbol{footnote}}

\newcommand{\citeSM}[1]{\cite{#1}}

\definecolor{chromeyellow}{rgb}{1.0, 0.65, 0.0}
\definecolor{darkpastelgreen}{rgb}{0.01, 0.75, 0.24}
\definecolor{darkolivegreen}{rgb}{0.33, 0.42, 0.18}
\definecolor{applegreen}{rgb}{0.55, 0.71, 0.0}
\definecolor{indigo}{rgb}{0.29, 0.0, 0.51}
\definecolor{orange}{rgb}{1,0.5,0}

\newcommand{\stvpatch}{\textcolor{red}{stv-patch}}
\newcommand{\tvpatch}{\textcolor{applegreen}{tv-patch}}
\newcommand{\svpatch}{\textcolor{chromeyellow}{sv-patch}}

\newcommand{\stvpatches}{\textcolor{red}{stv-patches}}
\newcommand{\tvpatches}{\textcolor{applegreen}{tv-patches}}
\newcommand{\svpatches}{\textcolor{chromeyellow}{sv-patches}}

\def\cvprPaperID{4873} 
\def\confYear{CVPR 2021}

\begin{document}

\title{Patch-VQ: `Patching Up' the Video Quality Problem}



\author{
	Zhenqiang Ying\textsuperscript{1\footnotemark[1]}, 
	Maniratnam Mandal\textsuperscript{1\footnotemark[1]}, 
	Deepti Ghadiyaram\textsuperscript{2\footnotemark[2]}, 
	Alan Bovik\textsuperscript{1\footnotemark[2]}\\
	\textsuperscript{1$\ddagger$}University of Texas at Austin, \textsuperscript{2}Facebook AI\\
	{\tt\small \{zqying, mmandal\}@utexas.edu, deeptigp@fb.com, bovik@ece.utexas.edu}
}

\maketitle

\begin{abstract}
\vspace{-1em}
No-reference (NR) perceptual video quality assessment (VQA) is a complex, unsolved, and important problem to social and streaming media applications. Efficient and accurate video quality predictors are needed to monitor and guide the processing of billions of shared, often imperfect, user-generated content (UGC). Unfortunately, current NR models are limited in their prediction capabilities on real-world, ``in-the-wild" UGC video data. To advance progress on this problem, we created the largest (by far) subjective video quality dataset, containing $39,000$ real-world distorted videos and $117,000$ space-time localized video patches (`v-patches'), and $5.5$M human perceptual quality annotations. Using this, we created two unique NR-VQA models: (a) a local-to-global region-based NR VQA architecture (called PVQ) that learns to predict global video quality and achieves state-of-the-art performance on $3$ UGC datasets, and (b) a first-of-a-kind space-time video quality mapping engine (called PVQ Mapper) that helps localize and visualize perceptual distortions in space and time. We will make the new database and prediction models available immediately following the review process.
\end{abstract}
\blfootnote{
\textsuperscript{$*\dagger$}Equal contribution}
\blfootnote{\textsuperscript{$\ddagger$} The entity that conducted all of the data collection/experimentation.
}
\vspace{-2em}


\section{Introduction}
User-generated content (UGC) and video streaming has exploded on social media platforms such as Facebook, Instagram, YouTube, and TikTok, each supporting millions and billions of users~\cite{tiktok}. It has been estimated that each day, about $4$ billion video views occur on Facebook~\cite{fbstats} and $1$ billion hours are viewed on YouTube \cite{ytstats}. Given the tremendous prevalence of Internet video, it would be of great value to measure and control the quality of UGC videos, both in capture devices and at social media sites where they are uploaded, encoded, processed, and analyzed.

Full-reference (FR) video quality assessment (VQA) models perceptually compare quality against pristine videos, while no-reference (NR) models involve no such comparison. Thus, NR video quality monitoring could transform the processing and interpretation of videos on smartphones, social media, telemedicine, surveillance, and vision-guided robotics, in ways that FR models are unable to. Unfortunately, measuring video quality without a pristine reference is very hard. Hence, though FR models are successfully deployed at the largest scales~\cite{ssim}, NR video quality prediction on UGC content remains largely unsolved, for several reasons.

\begin{figure}[t]
\begin{center}
\includegraphics[height=5.35cm,width=1\linewidth]{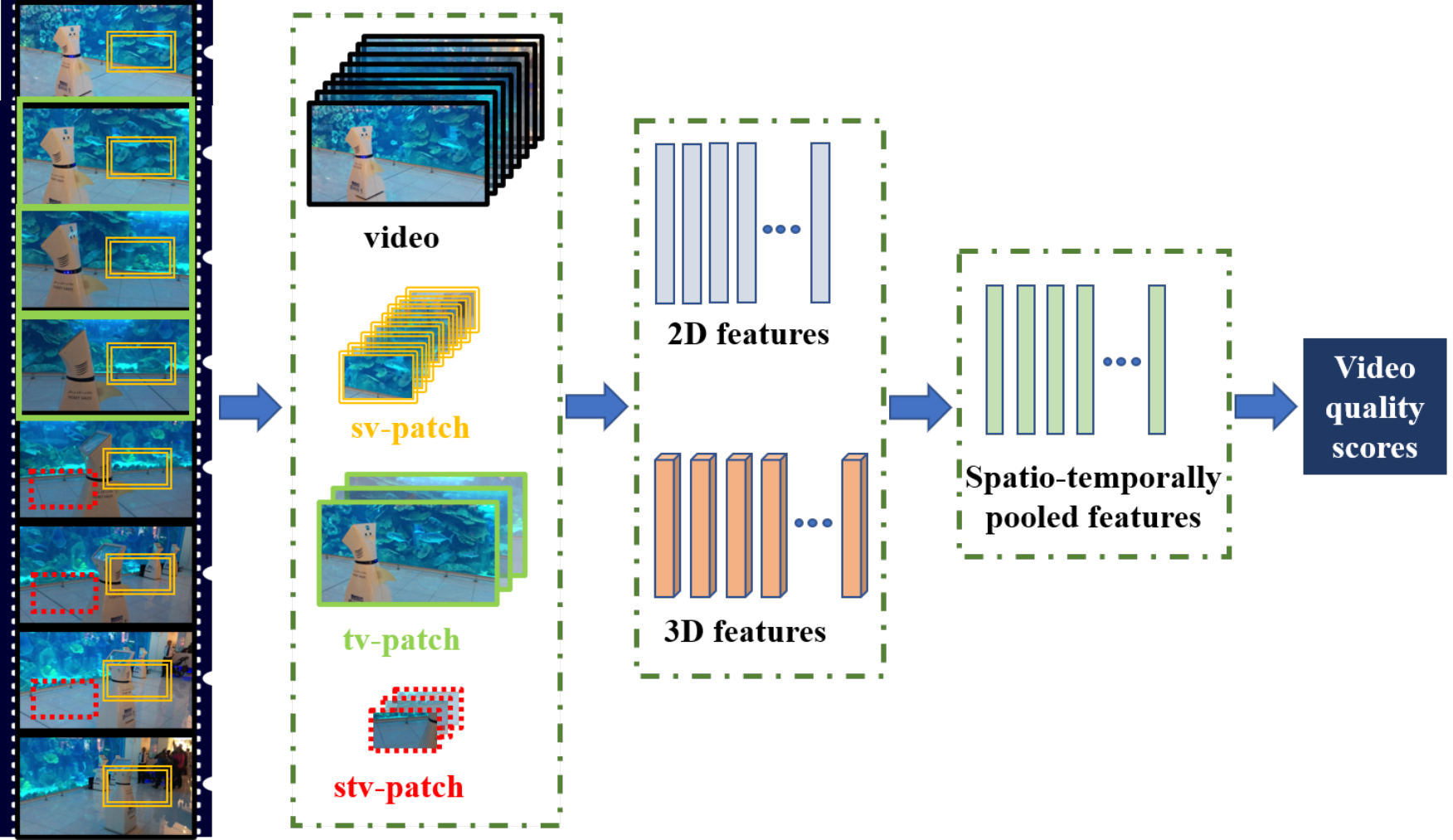}
\vspace{-1.5em}
\caption{\scriptsize{\textbf{Modeling local to global perceptual quality:} From each video, we extract three spatio-temporal video patches (Sec.~\ref{sec:dataset}), which along with their subjective scores, are fed to the proposed video quality model. By integrating spatial (2D) and spatio-temporal (3D) quality-sensitive features, our model learns spatial and temporal distortions, and can robustly predict both global and local quality, a temporal quality series, as well as space-time quality maps (Sec.~\ref{sec:qualityMaps}). Best viewed in color.}}
\vspace{-2.5em}
\label{fig:teaser}
\end{center}
\end{figure}

First, UGC video distortions arise from highly diverse capture conditions, unsteady hands of content creators, imperfect camera devices, processing and editing artifacts, frame rates, compression and transmission artifacts, and the way they are perceived by viewers. Inter-mixing of distortions is common, creating complex, composite distortions that are harder to model in videos. Moreover, it is well-known that the technical degree of distortion (e.g. amount of blur, blocking, or noise) does not correlate well with perceptual quality~\cite{mseLoveLeave}, because of neurophysiological processes that induce masking \cite{vpooling}. Indeed, equal amounts of distortions may very differently affect the quality of two different videos~\cite{jpegJSIT}.

Second, most existing video quality resources are too small and unrepresentative of the complex real-world distortions~\cite{epflPoli, livevqd, tum1080p, mcljcv, videoset, vqeg, csiq}. While three publicly available databases of authentically distorted UGC videos are available~\cite{konvid1k, livevqc, ytugc}, they are far too small to train modern, data-hungry deep neural networks. What is needed are very large databases of videos corrupted by real-world distortions, subjectively rated by large numbers of human viewers. However, conducting large-scale psychometric studies is much harder and time-consuming (per video) than standard object or action classification tasks.

Finally, although a few NR algorithms achieve reasonable performance on small databases~\cite{nrvqa4, nrvqa6, nrvqa7, nrvqa9, tlvqm, videval, vsfa}, most of them fail to account for the complex space-time distortions common to UGC videos. UGC distortions are often transient (e.g., frame drops, focus changes, and transmission glitches) and yet may significantly impact the overall perceived quality of a video~\cite{hysteresis}. Most existing models are frame-based, or use sample frame differences, and cannot capture diverse temporal impairments. 
\begin{table*}[t]
\captionsetup{font=scriptsize}
\vspace{-0.1in}
\scriptsize
\centering
\begin{tabular}{|c|c|c|c|c|c|c|c|c|c|}
\hline
Database & \thead{\scriptsize{\# Unique} \\ \scriptsize{contents}} & \thead{\scriptsize{\# Video} \\ \scriptsize{Duration (sec)}}   & \thead{\scriptsize{\# Distor-} \\ \scriptsize{tions}} & \thead{\scriptsize{\# Video} \\ \scriptsize{contents}} & \thead{\scriptsize{\# V-Patch} \\ \scriptsize{contents}} & Distortion type & \thead{\scriptsize{Subjective study} \\ \scriptsize{framework}} & \# Annotators &
\# Annotations  \\
\hline
MCL-JCV (2016) \cite{mcljcv} & 30 & 5 & 51 & 1,560 & 0 & Compression & In-lab & 150 & 78K \\

VideoSet (2017) \cite{videoset} & 220 & 5 & 51 & 45,760 & 0 & Compression & In-lab & 800 & - \\

UGC-VIDEO (2019) \cite{ugcvideo} & 50 & $>$ 10 & 10 & 550 & 0 & Compression & In-lab & 30 & 16.5K \\
\hline
CVD-2014 (2014) \cite{cvd2014} & 5 & 10-25 & - & 234 & 0 & In-capture & In-lab & 210 & - \\

LIVE-Qualcomm (2016) \cite{livequalcomm} & 54 & 15 & 6 & 208 & 0 & In-capture & In-lab & 39 & 8.1K \\
\hline
KoNViD-1k (2017) \cite{konvid1k} & 1,200 & 8 & - & 1,200 & 0 & In-the-wild & Crowdsourced & 642 & $\approx$ 205K \\

LIVE-VQC (2018) \cite{livevqc} & 585 & 10 & - & 585 & 0 & In-the-wild & Crowdsourced & 4,776 & 205K \\

YouTube-UGC (2019) \cite{ytugc} & 1,500 & 20 & - & 1,500 & 4,500 & In-the-wild & Crowdsourced & - & $\approx$ 600K \\


\hline
\textbf{Proposed database (LSVQ)} &  $39,075$ & 5-12 & - & $39,075$ & $117,225$ &  In-the-wild &  Crowdsourced & $6,284$ & $5,545,594$ \\
\hline
\end{tabular}
\caption{\scriptsize{\textbf{Summary of popular public-domain video quality datasets.} Legacy datasets contain singular synthetic distortions, whereas ``in-the-wild'' databases contain videos impaired by complex mixtures of diverse, real distortions.}} 
\vspace{-1em}
\label{tbl:datasets}
\end{table*}

We have made recent progress towards addressing these challenges, by learning to model the relationships that exist between local and global spatio-temporal distortions and perceptual quality. We built a large-scale public UGC video dataset of unprecedented size, comprising full videos and three kinds of spatio-temporal video patches (Fig.~\ref{fig:teaser}), and we conducted an online visual psychometric study to gather large numbers of human subjective quality scores on them. This unique data collection allowed us to successfully learn to exploit interactions between local and global video quality perception and to create algorithms that accurately predict video quality and space-time quality maps. We summarize our contributions below:

\vspace{-0.1in}
\begin{packed_enum}
\item \textbf{We built the largest video quality database in existence.} We sampled hundreds of thousands of open source Internet UGC digital videos to match the feature distributions of social media UGC videos. Our final collection includes $39,000$ real-world videos of diverse sizes, contents, and distortions, $26$ times larger than the most recent UGC dataset~\cite{ytugc}. We also extracted three types of v-patches from each video, yielding $117,000$ space-time video patches (``v-patches'') in total (Sec.~\ref{sec:dataset}).

\item \textbf{We conducted the largest subjective video quality study to date}. Our final dataset consists of a total of $5.5$M perceptual quality judgments on videos and v-patches from almost $6,300$ subjects, more than $9$ times larger than any prior UGC video quality study (Sec.~\ref{sec:human_study}).

\item \textbf{We created a state-of-the-art deep blind video quality predictor}, using a deep neural architecture that computes 2D video features using PaQ2PiQ~\cite{paq2piq}, in parallel with 3D features using ResNet3D~\cite{Hara_2017_ICCV}. The 2D and 3D features feed a time series regressor~\cite{inceptime} that learns to accurately predict both global video, as well as local space-time v-patch quality, by exploiting the relations between them. This new model, which we call Patch VQ (PVQ) achieves top performance on the new database as well as on smaller ``in-the-wild” databases~\cite{livevqc, konvid1k}, \emph{without finetuning} (Secs.~\ref{sec:baseline} and \ref{sec:cross_data}).

\item \textbf{We also create another unique prediction model that predicts first-of-a-kind space-time maps} of video quality by learning global-to-local quality relationships. This second model, called the PVQ Mapper, helps localize, visualize, and act on video distortions (Sec.~\ref{sec:qualityMaps}). 
\end{packed_enum}
\vspace{-0.5em}

\section{Related Work}\label{sec:background}
\noindent\textbf{Video Quality Datasets:} Several public legacy video quality datasets~\cite{epflPoli, livevqd, tum1080p, mcljcv, videoset, vqeg, csiq}
have been developed in the past decade. Each of these datasets comprise a small number of unique source videos (typically $10$-$15$), which are manually distorted by one of a few synthetic impairments (e.g., Gaussian blur, compression, and transmission artifacts). Hence, these datasets are quite limited in terms of content diversity and distortion complexity, and do not capture the complex characteristics of UGC videos. Early ``in-the-wild'' datasets~\cite{cvd2014, livequalcomm} included fewer than $100$ unique contents, while more recent ones such as KoNViD-1k~\cite{konvid1k}, LIVE-VQC~\cite{livevqc}, and YouTube-UGC~\cite{ytugc}  contain relatively more videos ($500$-$1500$ per dataset), yet insufficient to train deep models. A more recent dataset, FlickrVid-150k~\cite{mlspvqa} claims to contain a large number of videos, yet, has the following notable drawbacks: (a) Only $5$ quality ratings were collected on each video which, given the complexity of the task, are insufficient to compute reliable ground truth quality scores (at least $15$-$18$ is recommended \cite{itu_standard}). (b) the database is not publicly available, hence limiting its use for any experiments or to validate its statistical integrity. (c) the videos are all drawn from Flickr, which is largely populated by professional and advanced amateur photographers, hence is not representative of social media UGC content.

\noindent\textbf{Shallow NR VQA models:} 
Early NR VQA models were distortion specific \cite{nrvqa1, nrvqa3, nrvqa5, nrvqa8, nrvqa10, nrvqa12, nrvqa13} and focused mostly on transmission and compression related artifacts. More recent and widely-used NR image quality prediction algorithms have been applied to frame difference statistics to create space-time video distortion models~\cite{nrvqa4, nrvqa9, nrvqa14, nrvqa15}. In all these models, handcrafted statistical features are used to train shallow regression models to predict perceptual video quality, achieving high performance on legacy datasets. Recently proposed models~\cite{tlvqm, videval} use dozens or hundreds of such perceptually relevant features and achieve state-of-the-art performance on the leading UGC datasets, yet their predictive capability remains far below human performance.

\noindent \textbf{Deep NR VQA models:} There is more progress in the development of top-performing deep models for NR image quality prediction~\cite{ghadiyaram2014blind, deepConvIQA, bosseDeepIQA, fullyDeepIQA, nima, paq2piq, rankIQA, kedeMaIQA}, but relatively fewer deep NR-VQA models exist. The authors of~\cite{zhangvqa} proposed a general-purpose NR VQA framework based on weakly supervised learning and a resampling strategy. The NR VSFA \cite{vsfa} model uses a CNN to extract frame-wise features followed by a gated recurrent unit to capture temporal features. These, and other attempts~\cite{deepvqa, vmeon, zhangvqa, vsfa} mostly perform well on legacy datasets \cite{livevqd, mcljcv, csiq} and struggle on in-the-wild UGC datasets \cite{livevqc, ytugc, konvid1k}. MLSP-VQA~\cite{mlspvqa} 
reports high performance on~\cite{konvid1k}, but their code is not available, and we have been unable to reproduce their reported results.
\section{Large-Scale Dataset and Human Study}\label{sec:dataset_humanstudy}
Next, we present details of the newly constructed video quality dataset and the subjective quality study we conducted on it. The new database includes $39,075$ videos and $117,225$ ``v-patches" extracted from them, on which we collected about $5.5$M quality scores in total from around $6,300$ unique subjects. This new resource is significantly larger and more diverse than any legacy (synthetic distortion) databases \cite{livevqd, epflPoli, mcljcv, videoset} or in-the-wild crowd-sourced datasets \cite{konvid1k, livevqc, ytugc} ($26$ times larger than~\cite{ytugc}). We refer to the proposed dataset as the Large-Scale Social Video Quality (LSVQ) Database. 

\subsection{Building the Dataset} \label{sec:dataset}
\subsubsection{UGC-Like Data Collection and Sampling}
We selected two large public UGC video repositories to source our data: the Internet Archive (IA)~\cite{ia} and YFCC-100M~\cite{yfcc100m}, and collected a total of $400,000$ videos from them. Each video was randomly cropped to an average duration $7$ seconds\footnote{Cropping to a fixed duration was not possible, since a video must begin with a key frame to be decoded properly.} using \textit{ffmpeg} \cite{ffmpeg}.

\noindent\textbf{Sampling ``UGC-like'' videos:} Our dataset distinguishes itself from other in-the-wild video datasets in several ways. First, unlike KoNViD-1k~\cite{konvid1k}, we did not restrict the collected videos to have fixed resolutions or aspect ratios, making the proposed dataset much more representative of real-world content. Second, we did not apply scaling or further processing which could affect the quality of the content. Finally, to obtain ``UGC-like'' videos, we used a mixed integer programming method~\cite{mixedInteger} to match a set of UGC feature histograms. Specifically, we computed the following $26$ holistic spatial and temporal features on two video collections: (a) our aforementioned $400$K video collection from IA and YFCC-100M and (b) $19$K public, randomly selected videos from a social media website:


\begin{packed_enum}
\vspace{-0.7em}
\item \textit{Absolute Luminance} $L = R + G + B$. 
\item \textit{Colorfulness} using~\cite{measureColor}. 
\item \textit{RMS Luminance Contrast} \cite{peli}. 
\item Number of \textit{detected faces} using~\cite{faceDetection}.
\item \textit{Spatial Gaussian Derivative Filters} (3 scales, 2 orientations) from Leung-Malik filter bank~\cite{leungmalik}.
\item \textit{Temporal Gaussian Derivatives} (3 scales) first averaged along temporal dimension, followed by computing the mean and standard deviation along the spatial dimension.
\end{packed_enum}
\vspace{-0.7em}
The first five (spatial) features were computed on each frame, then the means and standard deviations of these features across all frames were obtained as the final features.


\begin{figure}[b]
\begin{center}
\includegraphics[ width=1\linewidth]{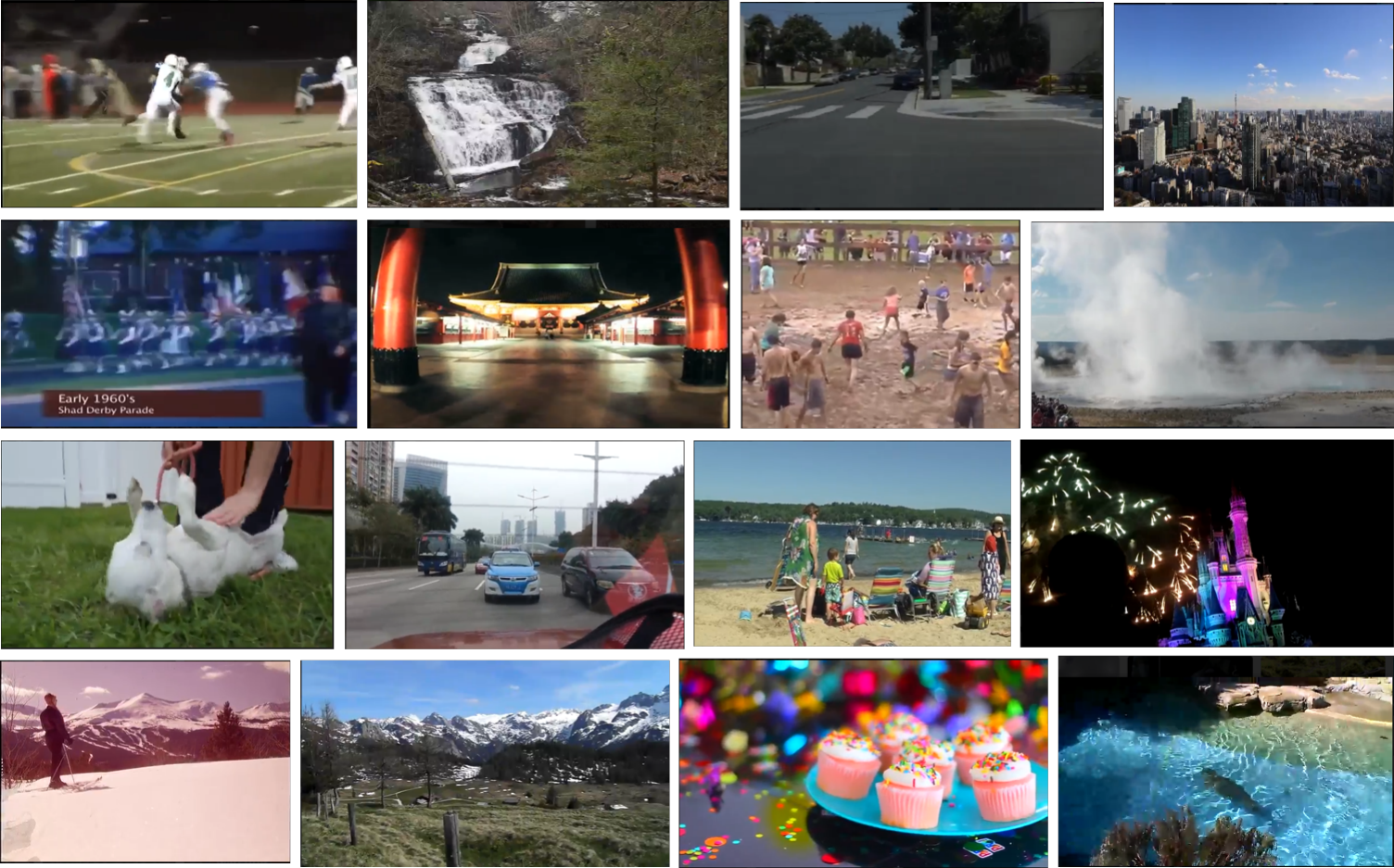}
\caption{\scriptsize{\textbf{Sample video frames from the new database}, each resized to fit. The actual videos are of highly diverse sizes and resolutions.}}
\vspace{-2em}
\label{fig:exemplarFLIVE}
\end{center}
\end{figure}

\begin{figure}[t]
\vspace{-1em}
\begin{center}
\includegraphics[width=1\linewidth]{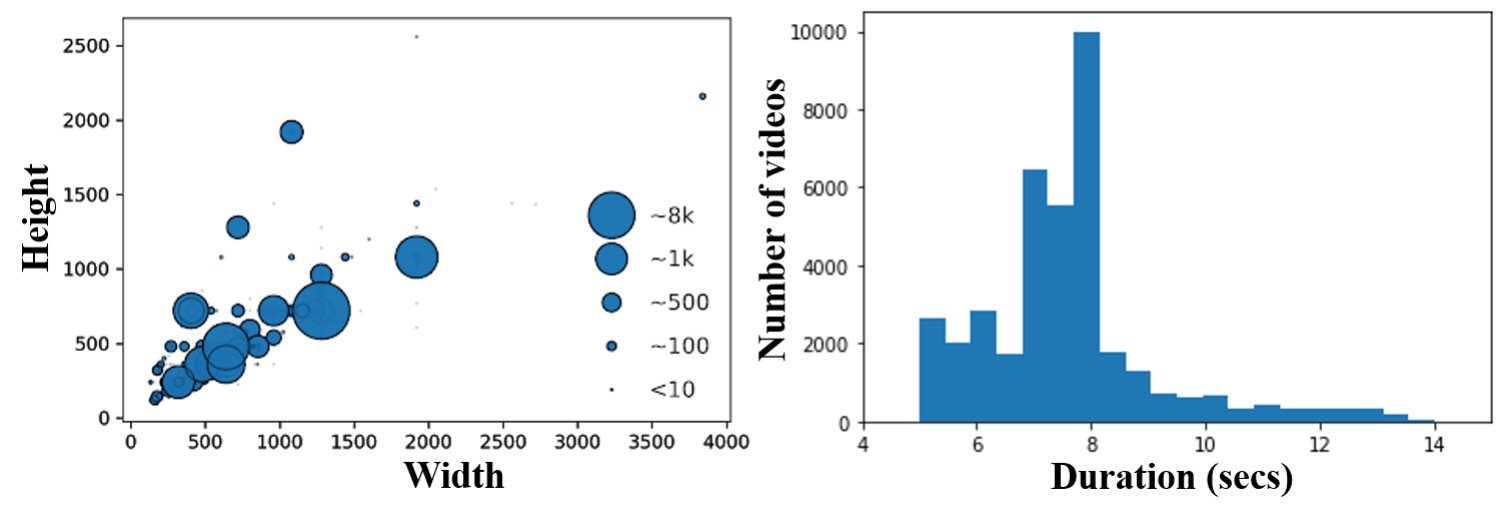} 
\vspace{-2em}
\caption{\scriptsize{\textbf{Left:} Scatter plot of video width versus video height with marker size indicating the number of videos having a given dimension in the new LSVQ database. \textbf{Right:} Histogram of the durations (in seconds) of the videos.}}
\vspace{-2.7em}
\label{fig:pixel_aspect}
\end{center}
\end{figure}

As mentioned, we sampled and matched feature histograms and in the end, arrived at about 39,000 videos, with roughly equal amounts from IA and YFCC-100M. Fig. \ref{fig:exemplarFLIVE} shows $16$ randomly selected video frames from LSVQ, while Fig.~\ref{fig:pixel_aspect} plots the diverse sizes, aspect ratios and durations of the final set of videos. It is evident that we obtained a diverse UGC video dataset that is representative in content, resolution, aspect ratios, and distortions. 

\vspace{-0.5em}

\subsubsection{Cropping Video-Patches}\label{sec:patch_cropping}
\begin{figure}[h]
\begin{center}
\includegraphics[width=1\linewidth]{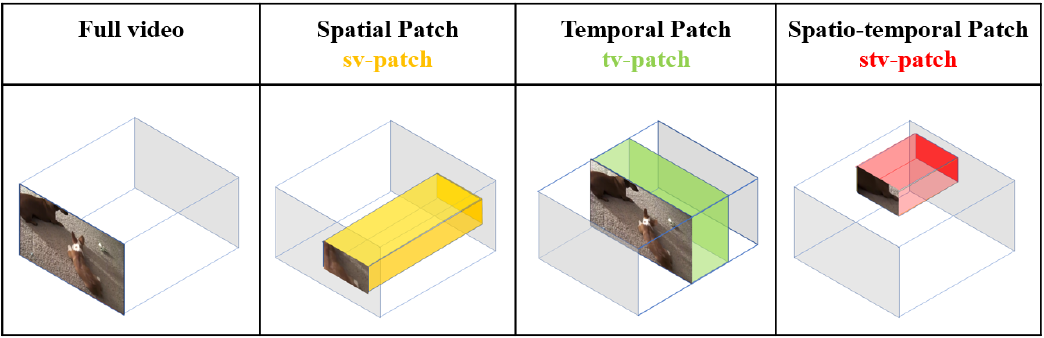}
\vspace{-2em}
\caption{\scriptsize{\textbf{Three kinds of video patches} (v-patches) cropped from random space-time volumes from each video in the dataset. All v-patches are videos.}}
\vspace{-0.3in}
\label{fig:egPatches}
\end{center}
\end{figure}

To closely study and model the relationship between global and local spatio-temporal qualities, we randomly cropped three different kinds of video patches or ``v-patches'' from each video: a spatial v-patch (\textbf{\svpatch}), a temporal v-patch (\textbf{\tvpatch}), and a spatio-temporal v-patch (\textbf{\stvpatch}). \underline{All three patches are videos} obtained by cropping an original video in space, time, or both space and time, respectively (Fig. \ref{fig:egPatches}). All v-patches have the same spatial aspect ratios as their source videos. Each \svpatch~has the same temporal duration as their source videos, but cropped to 40\% of spatial dimensions (16\% of area). Each \tvpatch~has the same spatial size as its source, but clipped to 40\% of temporal duration. Finally, each \stvpatch~was cropped to 40\% along all three dimensions. Every v-patch is entirely contained within its source, but the volumetric overlap of each \svpatch~and \tvpatch~with the same-source \stvpatch~did not exceed $25\%$ (suppl. material).

\begin{figure}[t]
\begin{center}
\includegraphics[width=\linewidth, trim={0em 0em 1.5em 0em},clip]{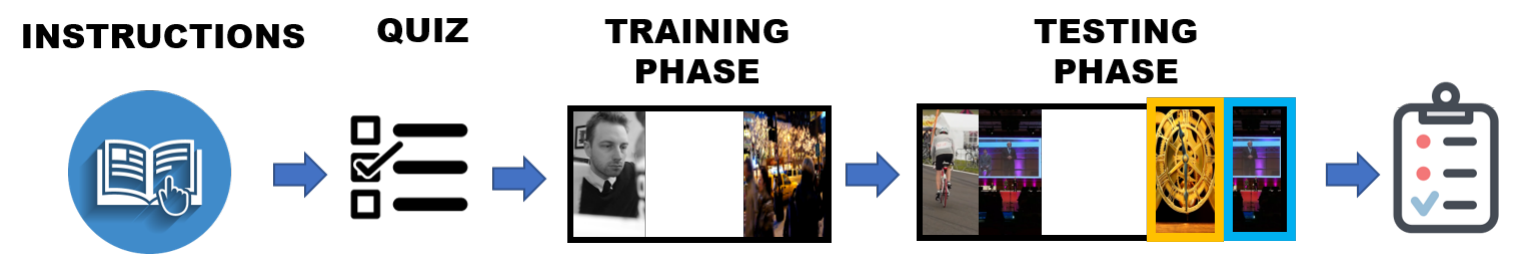}
\vspace{-2em}
\caption{\scriptsize{\textbf{Study workflow} for both video and v-patch sessions.}}
\vspace{-2em}
\label{fig:AMTdesign}
\end{center}
\end{figure}

\subsection{Subjective Quality Study}\label{sec:human_study}
Amazon Mechanical Turk (AMT) was used to collect human opinions on the videos and v-patches as in other studies~\cite{livevqc, ytugc, paq2piq, clive, koniq}. We launched two separate AMT tasks - one for videos and the other for the three video patches. A total of $6,284$ subjects were allowed to participate on both tasks. On average, we collected $35$ ratings on each video and v-patch. Subjects could participate in our study through desktops, laptops, or mobile devices.
\vspace{-0.5em}
\subsubsection{AMT Study Design}
The human intelligence task (HIT) pipeline is shown in Fig. \ref{fig:AMTdesign}. Each task began with general instructions, followed by a related quiz to check subjects' comprehension of the instructions, which they had to pass to proceed further. During training, each subject rated $5$ videos to become familiar with the interface and the task. Then, they entered the testing phase, in which they rated $90$ videos. Each video was played only once, following which the subject rated the video quality on a scale of $0$-$100$ by sliding a cursor along the rating bar (suppl. material). Subjects could report a video as inappropriate (violent or pornographic), static or incorrectly oriented. We ensured that each video was downloaded before playback to avoid rebuffering and stalling. At the end, each subject answered several survey questions about the study conditions and their demographics.
\vspace{-0.5em}
\subsubsection{Subject Rejection}
Next, we summarize the several checks we employed at various stages of the AMT task to identify and eliminate unreliable subjects~\cite{livevqc, clive} and participants with inadequate processing or network resources.

\noindent \textbf{During Instructions:} If a participant's browser window resolution, version, zoom, and the time taken to load videos did not meet our requirements (suppl. material), they were not allowed to proceed.

\noindent \textbf{During Training:} Although we ensured that each video was entirely downloaded prior to viewing, we also checked for any potential device-related video stalls. If the delay on any training video exceeded $2$ seconds, or the total delay over the five training videos exceeded $5$ seconds, the subject was not allowed to proceed (without prejudice). They were also stopped if a negative delay was detected (e.g., using plugins to speed up the video).

\noindent \textbf{During Task:} At the middle of each subject's task, we checked for instability of the internet connection, and if more than $50\%$ of the videos viewed until then had suffered from hardware stalls, the subject was disqualified. We also checked whether the subject had been giving similar quality scores to all videos, or was nudging the slider only slightly, both indicative of insincere ratings. 

\noindent \textbf{Post task: } In the test phase, of the $90$ videos, $4$, chosen at random, were repeated (seen twice at separate points), while another $4$ were ``golden" videos from KoNViD-1k~\cite{konvid1k}, for which subjective ratings were available. After each task, we rejected a subject if their scores on the same repeated videos or on the gold standard videos were not similar enough. 

Through all these careful checks, a total of 1,046 subjects were rejected over all sessions. 
\vspace{-0.5em}
\subsubsection{Data Cleaning}\label{sec:data_cleaning}
\vspace{-0.1em}
Following subject rejection, we conducted extensive data cleaning: \textbf{(1)} We excluded all scores provided by the subjects who were blocked, or for whom $>$ 50\% of the videos stalled during a session. \textbf{(2)} We removed ratings given by people who did not wear their prescribed lenses during the study ($1.13\%$), as uncorrected vision could affect perceived quality. \textbf{(3)} We applied ITU-R BT.500-14 \cite{itu_standard} (Annex 1, Sec 2.3) standard rejection to screen the remaining subjects. This resulted in $301$ subjects being rejected (about $2.6\%$). \textbf{(4)} To detect (and reject) outliers, we first calculated the kurtosis coefficient \cite{kurtosis} of each score distribution, to determine normality. We then applied the Z-score method in \cite{modzscore} if the distribution deemed Gaussian-like, and the Tukey IQR method \cite{tukey1977exploratory} otherwise (suppl. material). The total number of ratings collected after cleaning was around $5.6$M ($1.4$M on videos and $4.1$M on v-patches).

\vspace{-0.75em}
\subsubsection{Data Analysis}\label{sec:data_analysis}
\noindent \textbf{Inter-subject consistency:} On the cleaned data, we conducted an inter-subject consistency test~\cite{paq2piq, livevqc}. Specifically, we randomly divided the subjects into two equal and disjoint sets and computed the
{Spearman Rank Correlation Coefficient (\textbf{SRCC})}
\cite{kendall1948rank} between the two sets of MOS over $50$ such random splits. We achieved an average SRCC of \textbf{0.86} on full videos, and \textbf{0.71}, \textbf{0.71} and \textbf{0.67} for \svpatches, \tvpatches, and \stvpatches, respectively. This indicates a high degree of agreement between the human subjects, implying a successful screening process (suppl. material).

\noindent \textbf{Intra-subject consistency:} We computed the 
{Linear Correlation Coefficient (\textbf{LCC})
\cite{pcc}} between subjective MOS against the original scores on the ``golden" videos, obtaining a median PCC of \textbf{0.96} on full videos, and \textbf{0.946}, \textbf{0.95}, and \textbf{0.937} for \svpatches, \tvpatches, and \stvpatches, respectively. These high correlations further validate the efficacy of our data collection process.

\begin{figure}[t]
\begin{center}
\includegraphics[width=\linewidth]{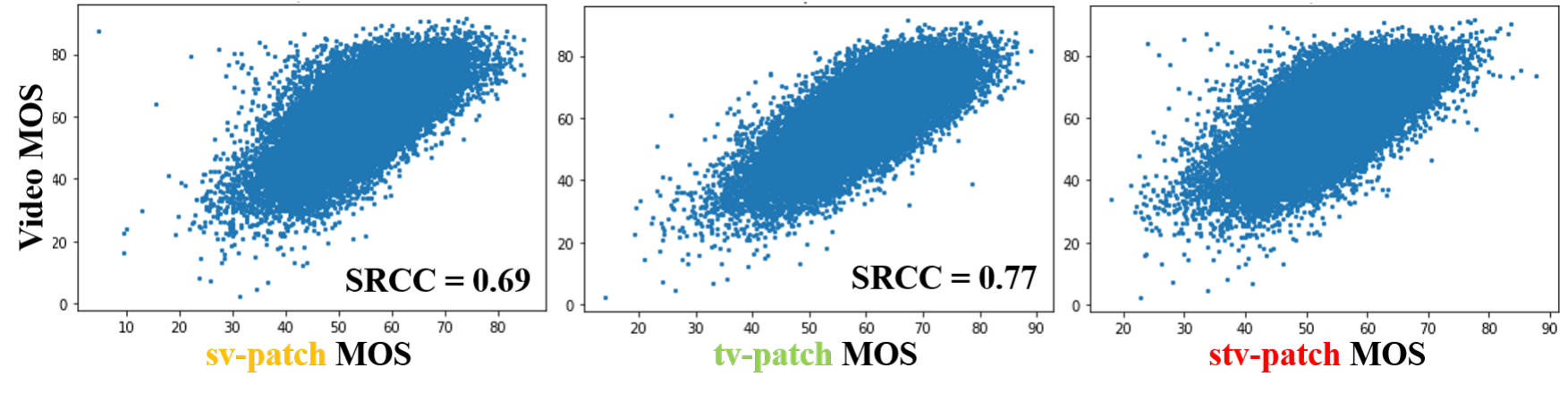}
\vspace{-2.2em}
\caption{\scriptsize{\textbf{Scatter plots of patch-video MOS correlations} Video MOS vs \svpatch ~(left), \tvpatch~(middle) and \stvpatch~(right) MOS cropped from the same video.}}
\label{fig:patchCorrel}
\vspace{-2.5em}
\end{center}
\end{figure}

\noindent\textbf{Relationship between patch and video quality:} Fig \ref{fig:patchCorrel} shows scatter plots of the video MOS against each type of v-patch MOS. The calculated SRCC between the video MOS and the \svpatch, \tvpatch~and \stvpatch~MOS was \textbf{0.69}, \textbf{0.77}, and \textbf{0.67} respectively, indicating strong relationships between global and local quality, even though the v-patches are relatively small volumes of the original video data. 

\begin{figure}[t]
\begin{center}
\includegraphics[width=\linewidth]{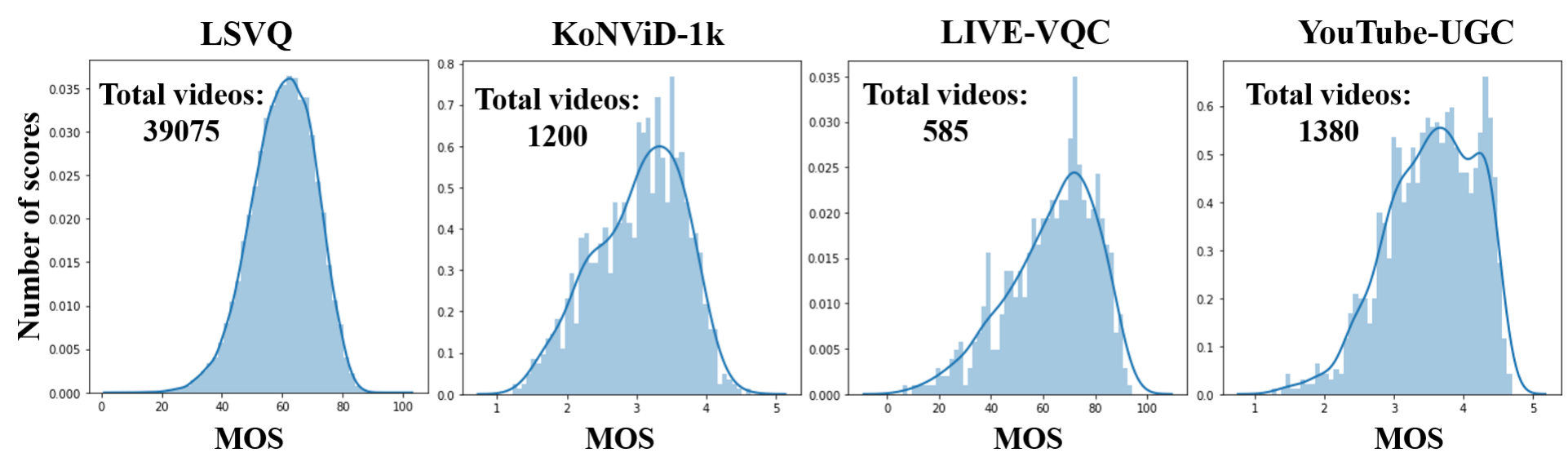}
\vspace{-2.2em}
\caption{\scriptsize{\textbf{Ground Truth MOS histograms of four ``in-the-wild” databases}. Starting from left, proposed LSVQ dataset, KoNViD-1k \cite{konvid1k}, LIVE-VQC \cite{livevqc}, and YouTube-UGC \cite{ytugc}.}}
\vspace{-2.5em}
\label{fig:MOShists}
\end{center}
\end{figure}

\noindent\textbf{MOS Distributions:} Fig. \ref{fig:MOShists} plots the MOS distribution of the videos in the new dataset as compared to other popular ``in-the-wild" video quality databases \cite{konvid1k, livevqc, ytugc}. The new dataset has a narrower distribution than the others, which again, matches actual social media data. Such a narrow distribution makes it more challenging to create predictive models that can parse finely differing levels of quality.

\section{Modeling a Blind Video Quality Predictor}\label{sec:modeling}
Taking advantage of the unique potential of the new dataset (Sec.~\ref{sec:dataset_humanstudy}), we created a deep video quality prediction model, which we refer to as Patch-VQ (PVQ), and a spatio-temporal quality mapper called PVQ-Mapper, both of which we describe next.
\subsection{Overview}\label{sec:baseline}
\begin{figure}[t]
\begin{center}
\includegraphics[height=0.33\textwidth]{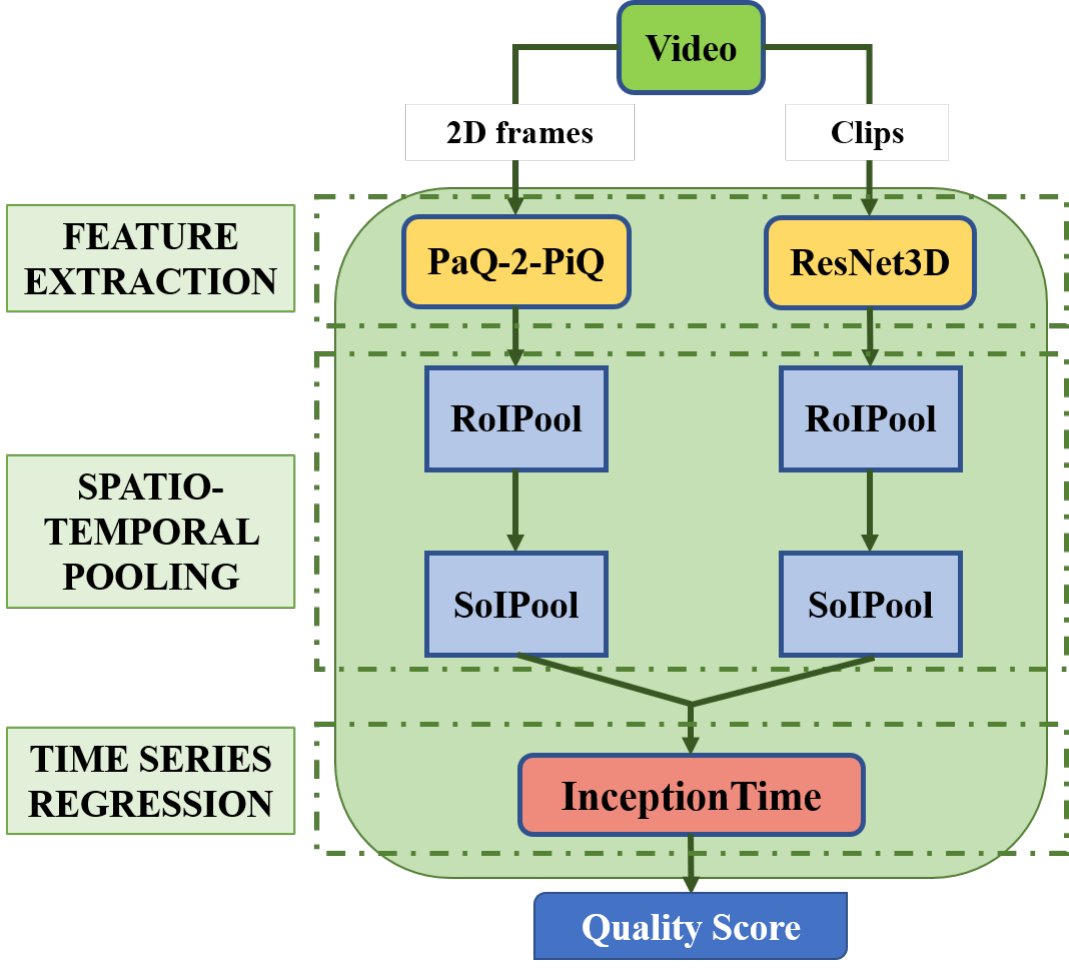}
\vspace{-1em}
\caption{\scriptsize{\textbf{Illustrating the proposed PVQ model} which involves 3 sequential steps: feature extraction, spatio-temporal pooling, and temporal regression (Sec.~\ref{sec:baseline}).}}
\vspace{-2.5em}
\label{fig:framework}
\end{center}
\end{figure}

Contrary to the way most deep image networks are trained, we did not crop, subsample, or otherwise process the input videos. Any such operation would introduce additional spatial and/or temporal artifacts, which can greatly affect video quality. Processing input videos of diverse aspect ratios, resolutions, and durations, however, makes training an end-to-end deep network impractical. To address this challenge, PVQ extracts spatial and temporal features on unprocessed original videos, and uses them to learn the local to global spatio-temporal quality relationships. As illustrated in Fig \ref{fig:framework}, PVQ involves three sequential steps: feature extraction, feature pooling, and quality regression. First, we extract features from both the 2D and 3D network streams, thereby capturing the spatial and temporal information from the whole video. Three kinds of v-patch features are also extracted from the output of both networks, using spatial and temporal pooling layers to capture local quality information. Finally, the pooled features from the video and the v-patches are processed by a time series network that effectively captures perceptual quality changes over time and predicts a single quality score per video. We provide more details of each step below.

\subsection{Feature Extraction}
To capture the spatial aspects of both perceptual video quality and frame content, we extracted per frame (2D) spatial features using the PaQ-2-PiQ backbone pre-trained on the LIVE-FB Dataset~\cite{paq2piq}. To capture temporal distortions, such as flicker, stutter, and focus changes, we extracted spatio-temporal (3D) features using a 3D ResNet-18~\cite{Hara_2017_ICCV} backbone, pre-trained on the Kinetics dataset~\cite{kinetics}.

\subsection{Feature Pooling}
Spatial and temporal pooling is applied in stages to extract features from the specified spatio-temporal regions of interest (v-patches), allowing us to model local-to-global space-time quality relationships.

\noindent\textbf{Spatial Pooling:} The extracted 2D and 3D features are independently passed through a spatial RoIPool (region-of-interest pooling) layer~\cite{fastRCNN, fasterRCNN}, with regions specified by the 3D v-patch coordinates. RoIPool helps compute a feature map with a fixed spatial extent of $2 \times 2$. The RoIPool layer generates 4 feature vectors of size $2048$ per frame and video clip, for all three v-patches and the full video.

\noindent\textbf{Temporal Pooling:} The RoIPool layer is followed by an SoIPool (segment-of-interest pooling) layer~\cite{soipool} that helps compute a feature map with a fixed temporal extent. Specifically, an SoIPool layer with a fixed temporal extent of $16$ is applied on both 2D and 3D features of each v-patch and the full video. The SoIPool layer yields $4$ feature vectors of size $16\times 2048$ per all three v-patches and the full video.

\subsection{Temporal Regression}
The resulting space-time quality features are fed to InceptionTime~\cite{inceptime}, a state-of-the-art deep model for Time Series Classification (TSC). InceptionTime consists of a series of inception modules (with intermittent residual connections) followed by a global average pooling and a fully connected layer. The inception modules learn changes in the quality features over time, which is crucial to accurately predict the global video quality. Although RNNs have been used to model temporal video quality~\cite{vsfa, youvqm}, we have found that InceptionTime~\cite{inceptime} is much faster and easier to train compared to RNN, does not suffer from vanishing gradients, and gives better performance.

\begin{table}[t]
\captionsetup{font=scriptsize}
\setlength\extrarowheight{1.0pt}
\centering
\footnotesize
\vspace{-0.8em}
\begin{tabular}{P{3cm}||P{0.85cm}|P{0.85cm}||P{0.85cm}|P{0.85cm}}
\hline
& \multicolumn{2}{c||}{\textbf{Test}} & \multicolumn{2}{c}{\textbf{Test-1080p}} \\
\hline
\hline
\textbf{Model} & \textbf{SRCC} & \textbf{LCC} & \textbf{SRCC} & \textbf{LCC} \\
\hline
\textit{BRISQUE} \cite{brisque} & 0.579 & 0.576 & 0.497 & 0.531 \\
TLVQM \cite{tlvqm} & 0.772 & 0.774 & 0.589 & 0.616  \\
VIDEVAL \cite{videval} & 0.794 & 0.783 & 0.545 & 0.554  \\
VSFA \cite{vsfa} & 0.801 & 0.796 & 0.675 & 0.704 \\
\hline
PVQ (w/o v-patch) & 0.814 & 0.816 & 0.686 & 0.708 \\
PVQ (w/ v-patch) & \textbf{0.827} & \textbf{0.828} & \textbf{0.711} & \textbf{0.739} \\
\hline
\end{tabular}
\caption{\footnotesize{\textbf{Performance on full-size videos in the LSVQ dataset.} Higher values indicate better performance. Picture based model is \textit{italicized.} 
}
}
\vspace{-1.5em}
\label{tbl:onLSVQ}
\end{table}

\section{Experiments}



\noindent\textbf{Train and test splits:} The entire dataset of videos, v-patches, and human annotations was divided into a training and \underline{two} test sets. We first selected those videos having both of their spatial dimensions greater than $720$, and reserved it for use as a secondary testing set (about $9\%$ of the LSVQ dataset: $3.5$K videos and $10.5$K v-patches). About $93.2$\% of the videos in the reserved set have resolutions $1080$p or higher, hence we will refer to  it as ``Test-1080p''. On the remaining videos, we applied a typical 80-20 split, yielding about $28.1$K videos (and $84.3$K v-patches) for training, and $7.4$K videos (and $22.2$K v-patches) for testing. 

\noindent \textbf{Input processing and training:} Each video was divided into $40$ clips of $16$ continuous frames. For feature extraction, we used a batch size of $8$ for 3D ResNet-18 and $128$ for PaQ-2-PiQ. For spatial and temporal pooling, we provide sets of spatio-temporal coordinates $(x_1, x_2, y_1, y_2, t_1, t_2)$ of each v-patch. When training InceptionTime, we used a batch size of $128$ and L1 loss to predict the output quality scores (details in suppl. material).

\noindent \textbf{Baselines and metrics:} The model comparisons were done on both videos and v-patches. We compared with a popular image model BRISQUE~\cite{brisque}, by extracting frame-level features and training an SVR and two other shallow NR VQA models, TLVQM~\cite{tlvqm} and VIDEVAL~\cite{videval}, that perform very well on existing UGC video databases. We also trained the VSFA~\cite{vsfa}, which extracts frame-level ResNet-50 \cite{resNet} features followed by a GRU layer to predict video quality. To study the efficacy of our local-to-global model, we trained two versions of our PVQ model, one with, and the other without the spatio-temporal v-patches. All models were trained and evaluated on the same train/test splits. Following the common practice in the field of video quality assessment, we report the performance using the correlation metrics SRCC and LCC.

\subsection{Predicting global video quality}
The quality prediction performance of the compared models on the new LSVQ dataset is summarized in Table \ref{tbl:onLSVQ}. As is evident, the shallow learner using traditional features (BRISQUE \cite{brisque}) did not perform well on our dataset. TLVQM \cite{tlvqm}, VSFA \cite{vsfa}, and VIDEVAL \cite{videval} performed better, indicating that they are capable of learning complex distortions. While both PVQ models (with and without patches) outperformed other models, including the v-patch data resulted in a performance boost on both test sets. Particularly on higher resolution test videos (Test-1080p), the proposed PVQ model (trained with v-patches) outperforms the strongest baseline by $\textbf{3.6\%}$ on SRCC.

\noindent \textbf{Performance on each v-patch:} Table~\ref{tbl:patches} sheds light into the capability of the compared models in predicting local quality. The two PVQ models delivered the best performance on all three types of v-patches, with the PVQ model trained on v-patches outperforming all baselines. From Tables~\ref{tbl:onLSVQ} and \ref{tbl:patches}, we may conclude that PVQ effectively captures global and different forms of local spatio-temporal video quality.  

\noindent \textbf{Contribution of 2D and 3D streams:} We also studied the contribution of the 2D and 3D features towards the performance of PVQ by training separate models on 2D (PVQ$_\text{2D}$) and 3D (PVQ$_\text{3D}$) features alone (Table \ref{tbl:ablation}). As can be observed, PVQ$_\text{3D}$ achieves higher performance than PVQ$_\text{2D}$ on both test sets. This further asserts that 3D features are more capable of capturing complex spatio-temporal distortions, and thus more favorable for VQA.

\noindent \textbf{Contribution of each v-patch:} To study the relative contributions of the three types of v-patches in PVQ, we trained three separate models utilizing each patch separately (Table \ref{tbl:ablation}). Among the three, we observe that the highest performance is achieved when trained on \stvpatches. Though \stvpatches~have relatively least volume (Fig.~\ref{fig:egPatches}), they contain the most localized information on video quality distortions, which could explain its better performance.

\begin{table}[t]
\captionsetup{font=scriptsize}
\setlength\extrarowheight{1.0pt}
\centering
\footnotesize
\vspace{-1em}
\begin{tabular}{P{2.2cm}||P{0.57cm}|P{0.57cm}||P{0.57cm}|P{0.57cm}||P{0.57cm}|P{0.57cm}}
\hline
& \multicolumn{2}{c||}{\textbf{\uppercase{\svpatch}}} & \multicolumn{2}{c||}{\textbf{\uppercase{\tvpatch}}} & \multicolumn{2}{c}{\textbf{\uppercase{\stvpatch}}} \\
\hline 
\hline
\textbf{Model} & \textbf{SRCC} & \textbf{LCC} & \textbf{SRCC} & \textbf{LCC} & \textbf{SRCC} & \textbf{LCC}\\
\hline
\textit{BRISQUE} \cite{brisque} & 0.469 & 0.417 &  0.465 & 0.485 & 0.476 & 0.462  \\
TLVQM \cite{tlvqm} & 0.575 & 0.543 & 0.523 & 0.536 & 0.561 & 0.563   \\
VIDEVAL \cite{videval} & 0.596 & 0.570 &  0.633 & 0.634  & 0.662 & 0.636   \\
VSFA \cite{vsfa} & 0.654 & 0.609 & 0.688 & 0.681 & {0.685} & 0.670  \\
\hline
PVQ (w/o v-patch) & {0.723} & {0.717} &	{0.696} & {0.701} & {0.651} & {0.643} \\
PVQ (w/ v-patch) & \textbf{0.737} & \textbf{0.720}	& \textbf{0.701} & \textbf{0.700}	& \textbf{0.711} & \textbf{0.707}  \\
\hline
\end{tabular}
\caption{\footnotesize{\textbf{Results on the three v-patches} in the LSVQ dataset. Picture based model is \textit{italicized.} 
}
}
\label{tbl:patches}
\vspace{-0.5em}
\end{table}


\begin{table}[t]
\captionsetup{font=scriptsize}
\setlength\extrarowheight{1.0pt}
\centering
\footnotesize
\begin{tabular}{P{3cm}||P{0.85cm}|P{0.85cm}|P{1.7cm}}
\hline
\textbf{Model} & \textbf{SRCC} & \textbf{LCC} & {\textbf{\# parameters}} \\
\hline
PVQ$_\text{2D}$ (w/ v-patch) & 0.774 & 0.774 & {16.3 M} \\
PVQ$_\text{3D}$ (w/ v-patch) & 0.805 & 0.805 & {38.3 M} \\
\hline
PVQ (w/ \svpatch) & 0.815 & 0.815 & {54.2 M} \\
PVQ (w/ \tvpatch) & 0.817 &	0.818 & {54.2 M} \\
PVQ (w/ \stvpatch) & 0.824 & 0.826 & {54.2 M} \\
\hline
{PVQ$_\text{Mobile}$} (w/ v-patch) & 0.774 & 0.779 & {10.9 M} \\
\hline
\end{tabular}
\caption{\footnotesize{\textbf{Ablation studies} conducted on the Test split of the LSVQ dataset.} Higher values indicate better performance. \vspace{-0.9em}}
\vspace{-1.5em}
\label{tbl:ablation}
\end{table}

\noindent \textbf{Mobile-friendly version:} 
We also implemented an efficient version of PVQ for mobile and embedded vision applications (PVQ$_\text{Mobile}$), using the 2D and 3D versions of MobileNetV2~\cite{sandler2018mobilenetv2, kopuklu2019resource} for the two branches, and by reducing the RoIPool output size to $1 \times 1$. Though there is a 6\% decrease in performance as compared to PVQ (w/ v-patch), our mobile model requires only $1/5$ as many parameters (Table \ref{tbl:ablation}) compared to PVQ (w/ v-patch) and $1/2$ as many parameters compared to VSFA~\cite{vsfa} ($24$M parameters).

\begin{figure}[t]
\begin{center}
\includegraphics[width=1\linewidth]{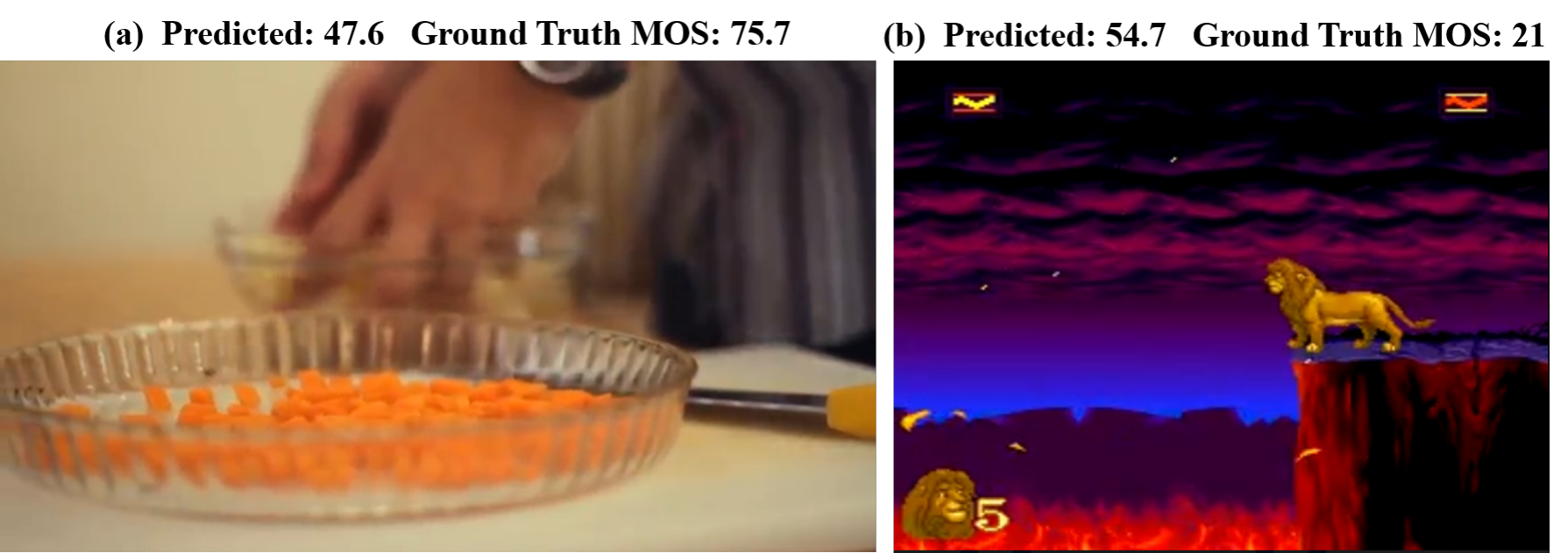}
\caption{\footnotesize{\textbf{Failure cases:} Frames from video examples where predictions differed the most from the human quality judgements.}}
\vspace{-0.3in}
\label{fig:failure_eg}
\end{center}
\end{figure}

\noindent \textbf{Failure cases:} The video in Fig \ref{fig:failure_eg} (a) was rated with a high score (MOS~=~$75.7$) by human subjects, but was underrated by PVQ (predicted MOS = $47.4$). We believe that an aesthetic ``bokeh" blur effect was interpreted as high quality content by subjects but such high levels of blur caused the model to predict low quality. The video in Fig \ref{fig:failure_eg} (b) was overrated by PVQ (predicted MOS = $54.7$), considerably higher than the subject rating (MOS = $21$). The video is of a computer generated game and does not appear very distorted. Yet, the subjects may have expected a higher resolution content for modern video games. These cases illustrate the challenges of creating models that closely align to human perception, while also highlighting the content diversity in the proposed dataset.

\subsection{Predicting perceptual quality maps} \label{sec:qualityMaps}
We adapted the PVQ model (Sec. \ref{sec:modeling}) to compute spatial and temporal quality maps on videos. Because of its flexible network architecture, PVQ is capable of predicting quality on any number (and sizes) of local spatio-temporal patches of an input video. We exploited this to create a temporal quality series and a first of its kind video quality map predictor, dubbed PVQ Mapper.

\noindent \textbf{Temporal quality series:} A video is uniformly divided into $16$ small temporal clips of $16$ (continuous) frames each\footnote{By changing the number of frames in each clip, the quality predictions can be made less or more dense.} and a single quality score per clip is computed, thus capturing a temporal series of perceptual qualities across a video.

\noindent \textbf{Space-time quality maps:} For space-time quality maps, we further divide each frame of each temporal clip defined above into a grid of $16 \times 16$ non-overlapping spatial blocks of the same aspect ratio as the frame and compute a local space-time video clip quality. Bi-linear interpolation was applied to spatially re-scale the spatio-temporal quality predictions to match the input dimensions. 

\begin{figure}[htb]
\vspace{-1em}
\begin{center}
    \includegraphics[width=0.9\linewidth]{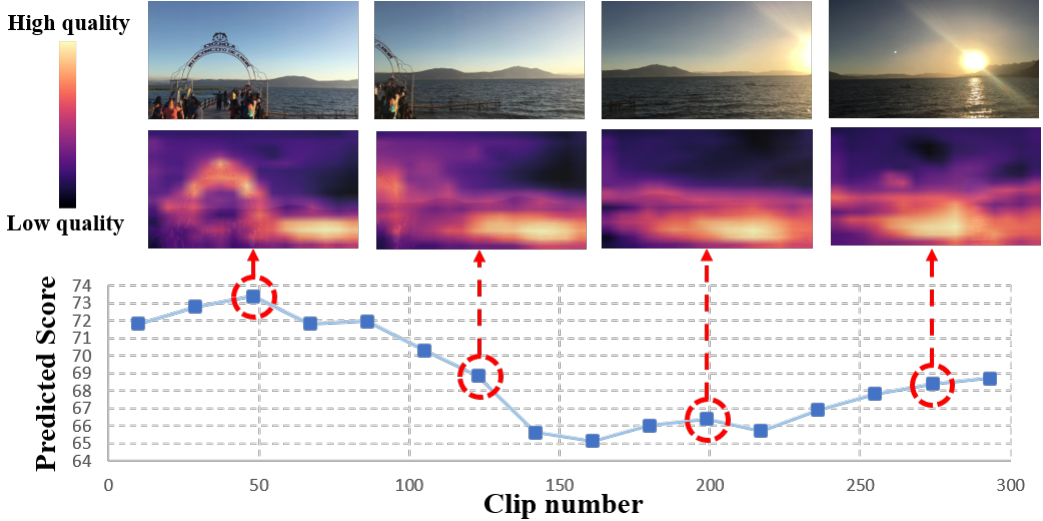}
    \vspace{-0.2cm}
    \captionof{figure}{\footnotesize{\textbf{Space-time quality maps:} Space-time quality maps generated on a video using the PVQ Mapper (Sec.~\ref{sec:qualityMaps}), and sampled in time for display. Four video frames are shown at top, with spatial quality maps (blended with the original frames using magma color) immediately under, while the bottom plots show the evolving quality of the video. Best viewed in color.}}
    \label{fig:space_time_quality}
\end{center}
\vspace{-2.5em}
\end{figure}

Fig.~\ref{fig:space_time_quality} depicts the temporal quality series and magma color space-time quality maps that were $\alpha$-blended ($\alpha=0.8$) with original frames picked from the center of each clip. The series shows the video quality evolving over time. As may be observed, PVQ Mapper was able to accurately capture local quality loss, distinguishing blurred and under-exposed areas from high-quality regions, and high-quality stationary backgrounds from fast-moving, streaky objects. 

\begin{figure}[h]
\begin{center}
    \includegraphics[width=1\linewidth,height=6.2cm]{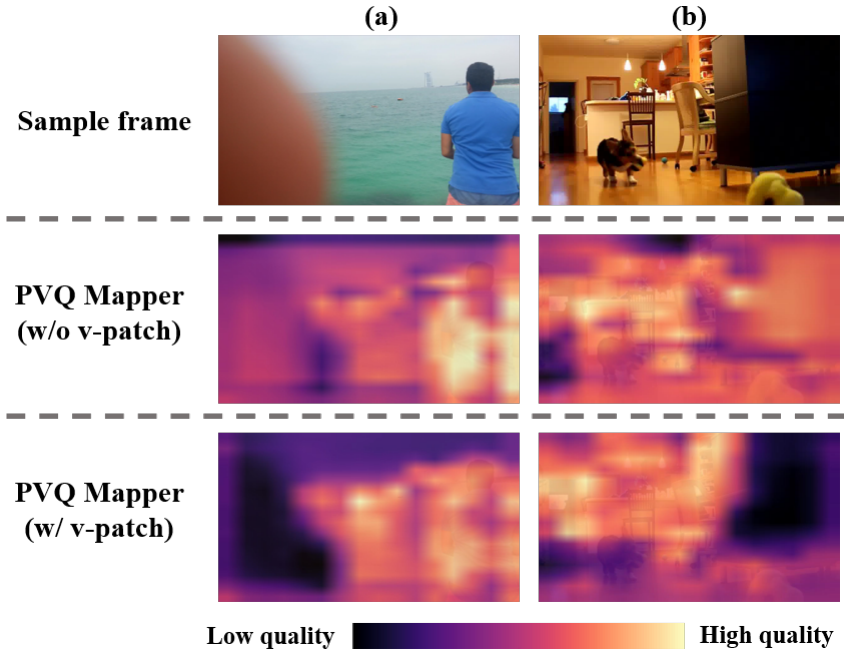}
    \captionof{figure}{\footnotesize{\textbf{Improvement in quality maps when PVQ-Mapper is trained with patches} illustrating that learning from both local space-time and global video quality yields more accurate predictions. Best viewed in color.}}
    \label{fig:spatial_quality}
    \vspace{-2em}
\end{center}
\end{figure}

\noindent\textbf{Do v-patches matter for quality maps?} Fig.~\ref{fig:spatial_quality} shows spatial quality maps on two sample videos generated by PVQ Mapper, trained with and without using v-patches. In Fig. \ref{fig:spatial_quality} (a), the object in the foreground is focus blurred, whereas in Fig. \ref{fig:spatial_quality} (b), the dog is motion blurred and the desk is underexposed. These local quality distortions are not captured with PVQ Mapper (w/o v-patch) as indicated in the middle row, but are distinctly evident in the output of PVQ Mapper (w/ v-patch) as indicated in the bottom row. This indicates that PVQ Mapper that uses v-patches is able to better learn from both global and local video quality features and human judgments of them, and hence predict more accurate quality maps.

\vspace{-1em}

\subsection{Cross-database comparisons} \label{sec:cross_data}
To emphasize the validity and generalizability of the PVQ model, we also tested it on the two popular, yet much smaller ``in-the-wild" video databases: KoNViD-1k~\cite{konvid1k} and LIVE-VQC~\cite{livevqc} (Table~\ref{tbl:datasets}). First, we compared the performance of PVQ against other popular models when each model was separately trained and tested on both datasets. As shown in Table \ref{tbl:comp_train_test}, PVQ competes very well with other models on KoNViD-1k, while improves the SRCC on LIVE-VQC by $\textbf{2.8\%}$ compared to the strongest baseline.

\begin{table}[t]
\captionsetup{font=scriptsize}
\setlength\extrarowheight{1.0pt}
\centering
\footnotesize
\vspace{-1em}
\begin{tabular}{P{3.1cm}||P{0.7cm}|P{0.7cm}||P{0.7cm}|P{0.7cm}}
\hline
& \multicolumn{2}{c||}{\textbf{KoNViD-1k} \cite{konvid1k}} & \multicolumn{2}{c}{\textbf{LIVE-VQC} \cite{livevqc}} \\
\hline
\textbf{Model} & \textbf{SRCC} & \textbf{LCC} & \textbf{SRCC} & \textbf{LCC} \\
\hline
\textit{BRISQUE} \cite{brisque} & 0.657 & 0.658 & 0.592 & 0.638 \\
V-BLIINDS \cite{bliinds2} & 0.710 & 0.704 & 0.694 & 0.718 \\
VSFA \cite{vsfa} & 0.773 & 0.775 & 0.773 & 0.795 \\
TLVQM \cite{tlvqm} & 0.773 & 0.769 & 0.799 & 0.803 \\
VIDEVAL \cite{videval} & 0.783 & 0.780 & 0.752 & 0.751 \\
\hline
PVQ (w/o v-patch) (Sec.~\ref{sec:modeling}) & \textbf{0.791} & \textbf{0.786} & \textbf{0.827}	& \textbf{0.837} \\
\hline 
\end{tabular}
\vspace{-0.5em}
\caption{\footnotesize{\textbf{Cross-database comparison 1:} Performance when all models are separately trained and tested on KoNViD-1k \cite{konvid1k} and LIVE-VQC \cite{livevqc}.}} 
\label{tbl:comp_train_test}
\end{table}

To further study the generalizability of PVQ, we also compared the performance of all models when trained on the proposed dataset (LSVQ) but tested on the two aforementioned datasets. From Table \ref{tbl:comp_test}, it may be seen that PVQ transferred very well to both datasets. Specifically, our model outperforms the strongest baseline by $\textbf{0.7\%}$ and $\textbf{3.6\%}$ boost in SRCC on KoNViD-1k and LIVE-VQC respectively. This degree of database independence, both highlights the representativeness of the new LSVQ dataset and the general efficacy of the proposed PVQ model.

\begin{table}[t]
\captionsetup{font=scriptsize}
\setlength\extrarowheight{1.0pt}
\centering
\footnotesize
\begin{tabular}{P{3.1cm}||P{0.7cm}|P{0.7cm}||P{0.7cm}|P{0.7cm}}
\hline
& \multicolumn{2}{c||}{\textbf{KoNViD-1k} \cite{konvid1k}} & \multicolumn{2}{c}{\textbf{LIVE-VQC} \cite{livevqc}} \\
\hline
\textbf{Model} & \textbf{SRCC} & \textbf{LCC} & \textbf{SRCC} & \textbf{LCC} \\
\hline
\textit{BRISQUE} \cite{brisque} & 0.646 & 0.647 & 0.524 & 0.536 \\
TLVQM \cite{tlvqm} & 0.732 & 0.724 & 0.670	& 0.691 \\
VIDEVAL \cite{videval} & 0.751 & 0.741 & 0.630	& 0.640 \\
VSFA \cite{vsfa} & 0.784 & 0.794 & 0.734 & 0.772 \\
\hline
PVQ (w/o v-patch) (Sec. \ref{sec:modeling}) & {0.782} & {0.781} &	{0.747} & {0.776} \\
PVQ (w/ v-patch) (Sec. \ref{sec:modeling}) & \textbf{0.791} & \textbf{0.795} &	\textbf{0.770} & \textbf{0.807} \\
\hline 
\end{tabular}
\caption{\footnotesize{\textbf{Cross-database comparison 2:} Performance when all models are separately trained on the new LSVQ database, then evaluated on KoNViD-1k \cite{konvid1k} and LIVE-VQC \cite{livevqc} \textbf{without fine-tuning.}\vspace{-1.2em}}} 
\vspace{-1.5em}
\label{tbl:comp_test}
\end{table}

\section{Concluding Remarks}\label{sec:conclusion}
Predicting perceptual video quality is a long-standing problem in vision science, and more recently, deep learning. In recent years, it has dramatically increased in importance along with tremendous advances in video capture, sharing and streaming. Accurate and efficient video quality prediction demands the tools of large-scale data collection, visual psychometrics, and deep learning. To progress towards that goal, we built a new video quality database, which is substantially larger and diverse than previous ones. The database contains patch-level annotations that enable us (and others) to make global-to-local and local-to-global quality inferences, culminating in the accurate and generalizable PVQ model. We also created a space-time video quality mapping model, called PVQ Mapper, which utilizes learned patch quality attributes to accurately infer local space-time video quality, and is able to generate accurate spatio-temporal quality maps. We believe that the new LSVQ dataset, the PVQ model, and PVQ Mapper, can significantly advance progress on the UGC VQA problem, and enable quality-based monitoring, ingestion, and control of billions of videos streamed on social media platforms.
\clearpage
{\small
\bibliographystyle{unsrt}
\bibliographystyle{ieee}
\bibliography{main}
}
\pagenumbering{gobble}
\newpage
\newpage
\onecolumn
\setcounter{figure}{0}
\begingroup
\renewcommand\thesubsection{\Alph{subsection}}
\def\thesection{\Alph{section}}
\part*{Supplementary Material -- \\ Patch-VQ: `Patching Up' the Video Quality Problem}
\subsection{Cropping Patches}
\noindent \textbf{Deciding number of scales for cropping v-patches: } In a psychometric study, specifically based on evaluating video quality, a subject needs roughly 15-20 seconds to rate each content. This limited the number of v-patches we could collect ratings on, and thus we decided to only include \textbf{one scale} for each type of v-patches. Scale here defines the dimensions of the v-patches, or the proportion of the video data contained in the patches. For simplicity, we use the same scale (40\% of original dimensions) for extracting the three types of v-patches. Additional examples of extracted v-patch triplets have been shown in Fig. \ref{fig:suppl_Patches}. 

\noindent \textbf{Deciding size of v-patches: } Empirically, \svpatches~cropped at large scales are not local enough, and do not capture the local quality features satisfactorily. Alternately, smaller scales result in \tvpatches~too short in duration to collect reliable judgements. Similarly, the resulting \stvpatches~are too small and short to rate comprehensibly and reliably. We determined 40\% to be the most suitable scale after examining v-patch samples.

\begin{figure}[h]
\begin{center}
\includegraphics[width=0.7\linewidth]{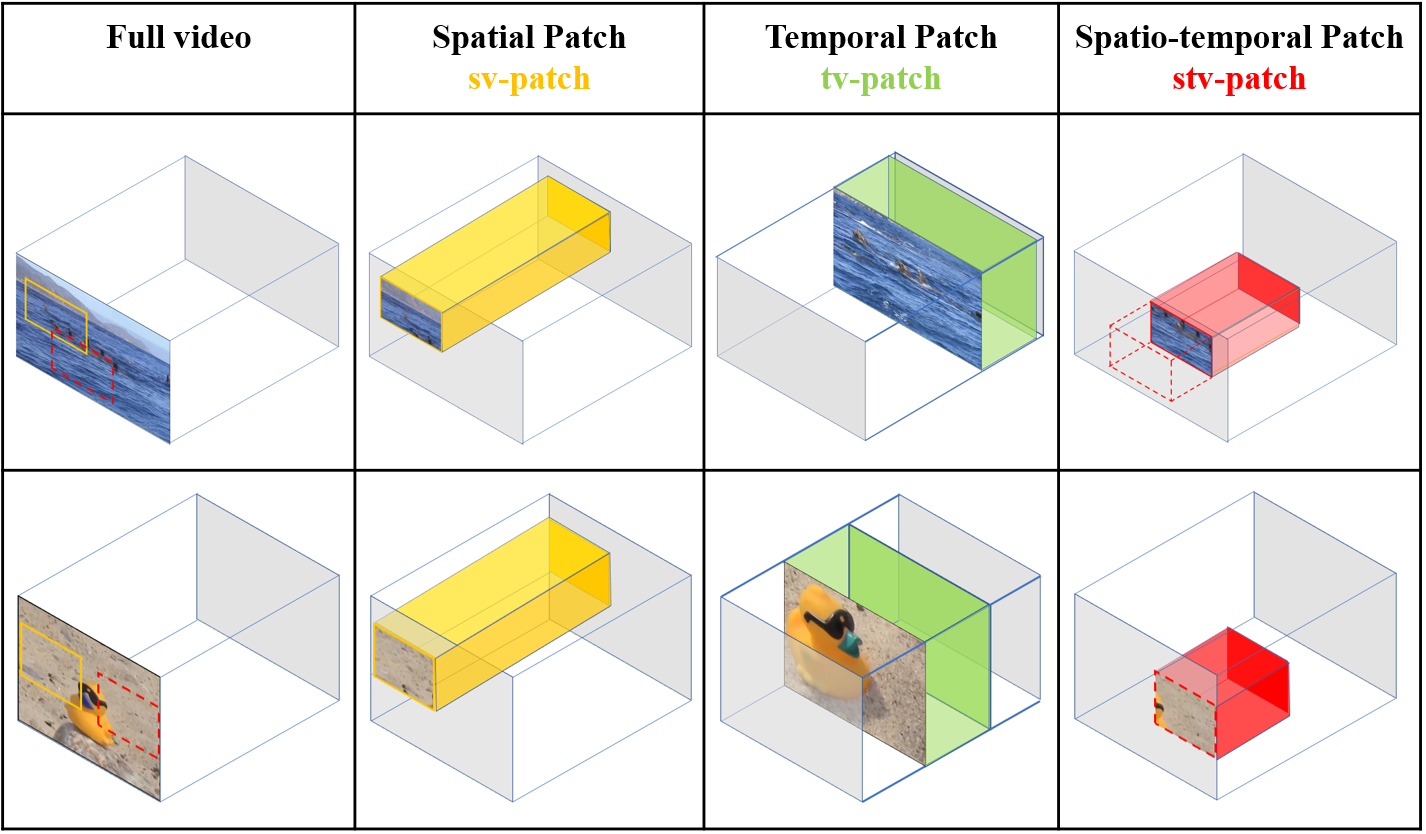}
\caption{\scriptsize{\textbf{Examples of video patch} (v-patch) triplets cropped from random space-time volumes from two exemplar videos in the dataset. All v-patches are videos.}}
\vspace{-0.3in}
\label{fig:suppl_Patches}
\end{center}
\end{figure}


\subsection{Dataset}


\subsubsection{Inter-subject consistency plots:} \label{sec:inter-cons}

We have mentioned the average SRCC values, representative of the inter-subject consistencies, in Sec. \ref{sec:data_analysis}. Along with that, we present the scatter plots of the two sets of subject MOS in Fig. \ref{fig:inter-consistency}. The narrow spread of the plots shows the high agreement, and hence higher consistency, among subject ratings. We also notice that the spread is highest (or, the correlation is lowest) in the case of \stvpatches. This can be attributed to the fact that they account for only $6.4\%$ of the video pixel volume, and sometimes, distortions prominent locally, might get masked or have little impact on perceived global video quality.
\begin{figure}[hbt]
\begin{center}
\includegraphics[width=1\linewidth]{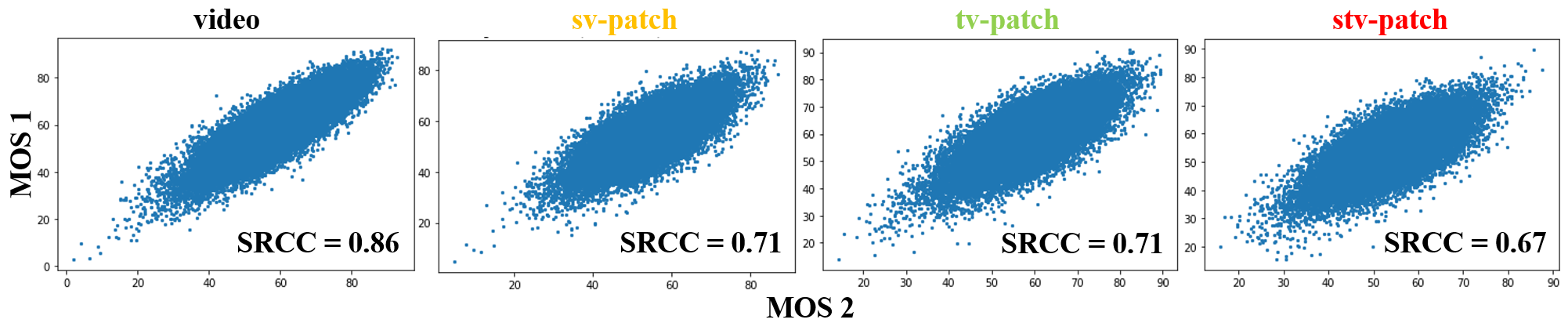}
\vspace{-2em}
\caption{\scriptsize{\textbf{Inter-subject consistency: } Inter-subject scatter plot of MOS calculated between random $50\%$ divisions of the human labels on all $39$K videos (first from left) into disjoint subject sets. The same is plotted for the \svpatches~(second), \tvpatches~(third) and \stvpatches~(fourth).
}}
\vspace{-2em}
\label{fig:inter-consistency}
\end{center}
\end{figure}

\subsubsection{Consistency among subject demographics: }
We utilized the subject data to study the effects of device parameters on MOS. The SRCC calculated between laptops and desktop computers (the most used devices in the study) was $\textbf{0.7}$, whereas that between videos viewed on phones and other devices was $\textbf{0.5}$. Although we collected relatively little data ($3.7\%$) from phones, this reinforces the notion that perceptual video quality is impacted by viewing on a small device screen. We obtained the following correlations between the two major resolutions: $768\times 1366$ and $640\times 360$ ($\textbf{0.76}$); major viewing distances: less than $15$ inches and $15$-$30$ inches ($\textbf{0.76}$); major age groups: $20$-$30$ and $30$-$40$ ($\textbf{0.79}$); and genders ($\textbf{0.8}$), all of which are high, but low enough to be suggestive of further study. The consistency among the ratings from diverse subject demographics, when accumulated, result in the overall high consistency of the data (Sec. \ref{sec:inter-cons}), validating our data collection and cleaning methodologies. 

\subsubsection{Effect of playback delays on video quality:}
Delays during playback could impact video quality \citeSM{livevqc}. We found that $>$ 96\% of the videos were viewed with delays $<$ 1s., while 86\% of the videos played without delays. By comparing the scores of the delayed videos against the ``golden" scores, we found that device delays had negligible impact on the mean scores, and that eliminating scores associated with delays did not impact data consistency. Hence, we did not impose device delays as a rejection criteria. 

\subsubsection{Outlier rejection:}
We removed the outliers in our data in two steps as described briefly in Sec. \ref{sec:data_cleaning} - \textbf{outlier subject rejection} and \textbf{outlier score rejection}. The former rejection was video independent, whereas the latter was subject independent. Here, we elaborate the outlier score rejection, which was executed on all videos individually.
We followed the standard outlier rejection techniques, but the technique applied was dependent on the score distribution. If, for a video, the scores were (approximately) Gaussian, the modified Z-scores method \citeSM{modzscore} was applied, which is based on calculating the standard deviation of the distribution. Calculating the kurtosis helped determine the normality of the score distribution. Alternately, if the scores were deemed to be not normal, then we applied the Tukey IQR \citeSM{tukey1977exploratory} detection technique, which is based on calculating the interquartile range and is a more generalized method. Tuning the outlier rejection methods based on the nature of the score distribution yielded better consistency scores.  

\subsection{Modeling Details}
\noindent For training PVQ (Sec. \ref{sec:modeling}), we used the Adam optimizer with $\beta_1=.9$ and $\beta_2=.99$, a weight decay of $.01$.
The initial learning rate was set to be 0.001 and we followed the 1cycle policy \citeSM{onecycle} to adjust the learning rate on the fly. We trained each model for $10$ epochs and report the performance of the model on the two testing sets. 

\subsection{Amazon Mechanical Turk (AMT) Study}

\subsubsection{Study Requirements: }\label{sec:study_requirements}
Each video batch (and thus each video) was published on AMT in four phases. 
The first two phases targeted ``reliable" workers (with AMT ratings $>$ 95\%, and $>$ 10,000 HITs), who helped eliminate inappropriate (violent or pornographic) content and static videos. In the latter two phases, we reduced the numbers to 75\% and 1000, respectively. 

As each subject was viewing the instructions, we monitored several parameters to ensure that they could effectively participate. The following eligibility criteria were imposed - 
\begin{packed_enum}
\item \textbf{Browser Window Resolution: } At least 480p for mobile devices and 720p for others.
\item \textbf{Browser Zoom: } Set to 100\%.
\item \textbf{Browsers: } Latest versions of Chrome, Firefox, Edge, Safari, and Chrome.
\item \textbf{Loading Time: } Must be less than 20 secs for all the training videos.
\end{packed_enum}
In case they failed to meet any of the above criterion, subjects were prevented from progressing and informed accordingly. Apart from these, the subjects were also required to take a quiz reflecting their understanding of the instructions, and were allowed to proceed only if they answered at least five out of the six questions.

\subsubsection{Interface: }
The AMT interface comprised of a series of instruction pages, followed by the quiz, before they could start rating the videos. Workers were allowed to view the introductory page (Fig. \ref{fig:interface1}) before accepting to participate in the study. If accepted, they had to go through the instruction pages (Fig. \ref{fig:interface2}, \ref{fig:interface3}, \ref{fig:interface4}, \ref{fig:interface5}), which were timed. During the instructions, we checked whether they satisfied the study criteria as described in Sec. \ref{sec:study_requirements}. Following the instruction pages, they had to pass the quiz (Fig. \ref{fig:interface_quiz}) in order to proceed to the training and testing phases. The task included rating the played video (Fig. \ref{fig:interface_vid}) on a Likert scale \citeSM{likert} marked with \textit{BAD}, \textit{POOR}, \textit{FAIR}, \textit{GOOD}, and \textit{EXCELLENT}, as demonstrated in Fig. \ref{fig:interface_slider}. A similar interface was used for the v-patch sessions as well.

\begin{figure}[hbt]
\begin{center}
\includegraphics[width=0.74\linewidth]{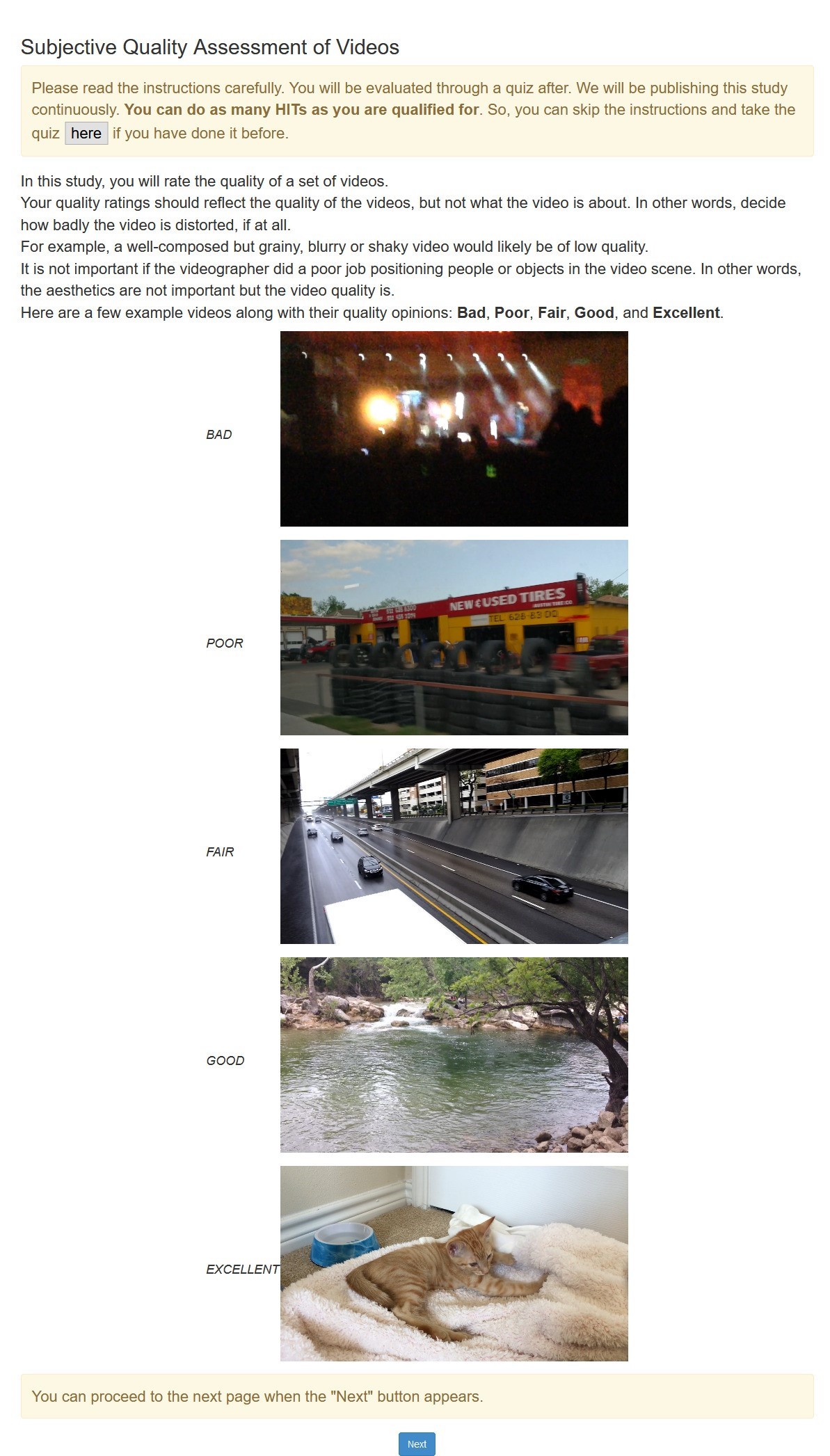}
\caption{\scriptsize{\textbf{Introductory Page} }}
\vspace{-2em}
\label{fig:interface1}
\end{center}
\end{figure}

\begin{figure}[hbt]
\begin{center}
\includegraphics[width=1\linewidth]{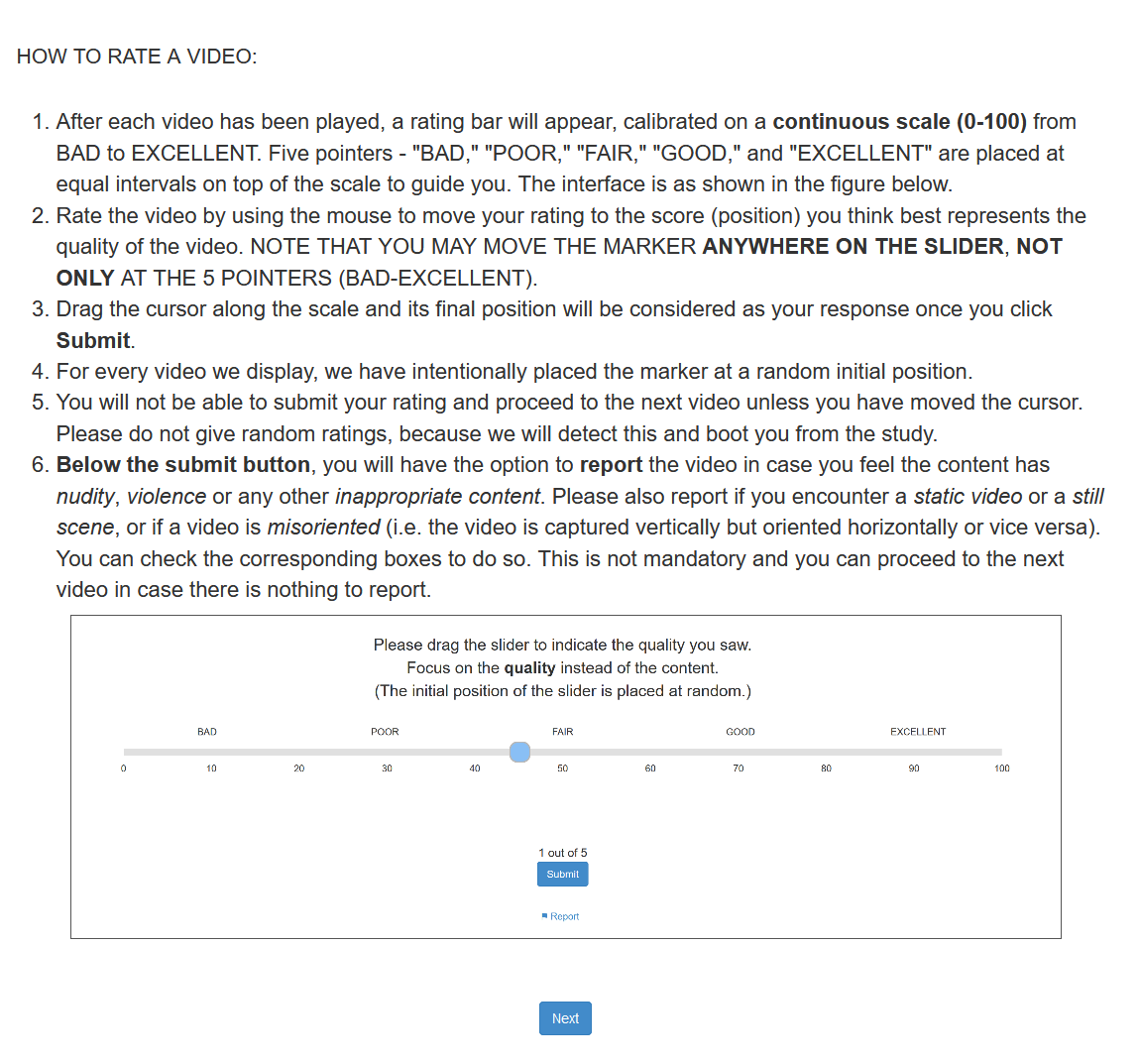}
\vspace{-1em}
\caption{\scriptsize{\textbf{Instruction Page 1} }}
\vspace{-2em}
\label{fig:interface2}
\end{center}
\end{figure}

\begin{figure}[hbt]
\begin{center}
\includegraphics[width=1\linewidth]{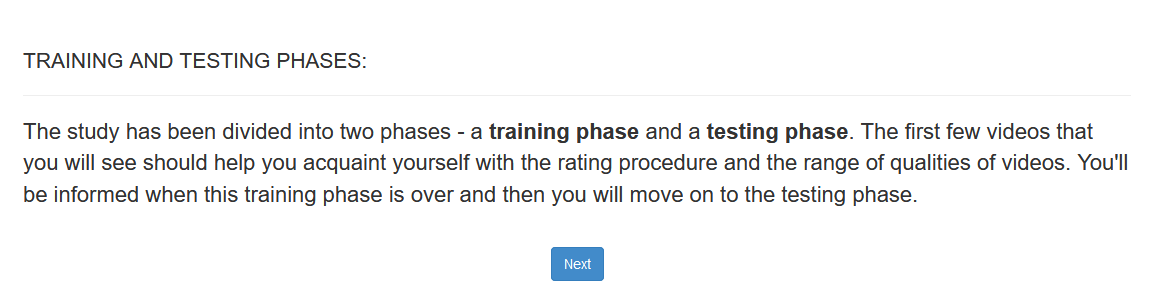}
\vspace{-1em}
\caption{\scriptsize{\textbf{Instruction Page 2} }}
\vspace{-2em}
\label{fig:interface3}
\end{center}
\end{figure}

\begin{figure}[hbt]
\begin{center}
\includegraphics[width=1\linewidth]{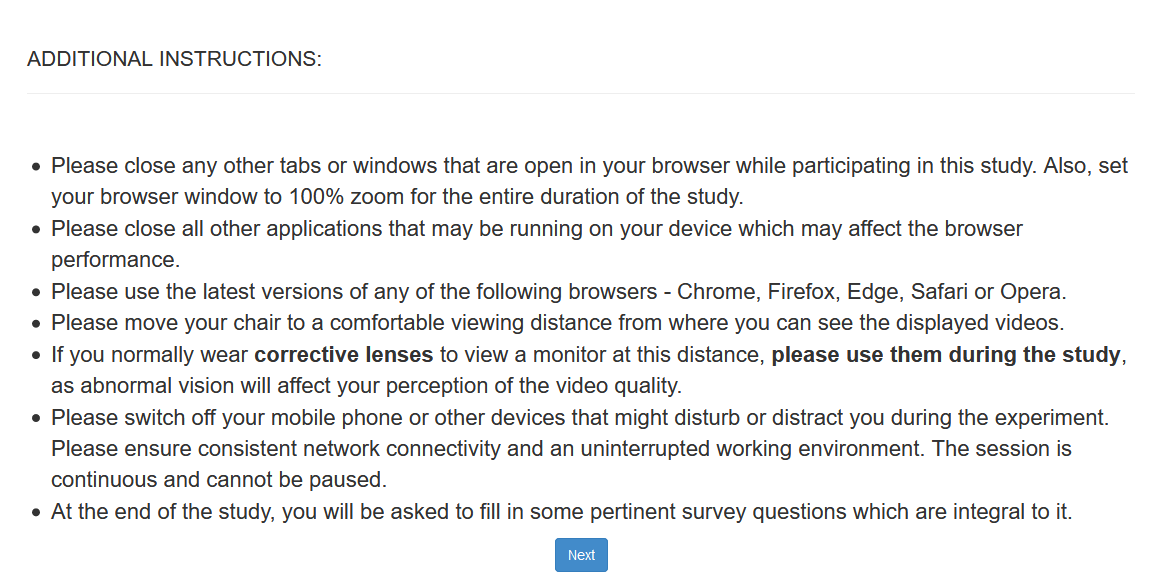}
\vspace{-1em}
\caption{\scriptsize{\textbf{Instruction Page 3} }}
\vspace{-2em}
\label{fig:interface4}
\end{center}
\end{figure}

\begin{figure}[hbt]
\begin{center}
\includegraphics[width=1\linewidth]{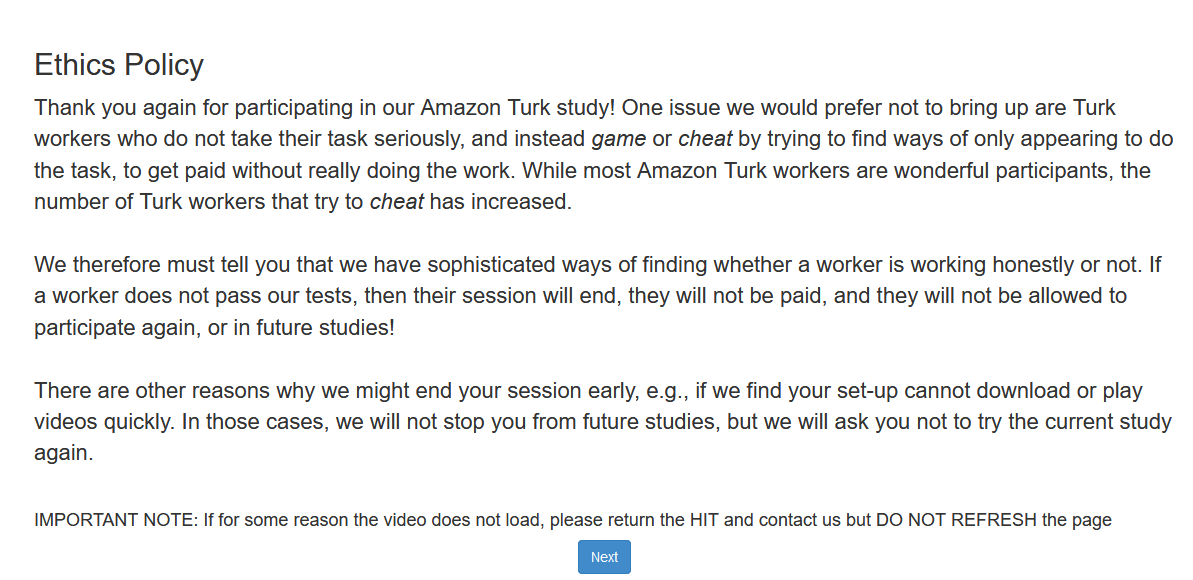}
\vspace{-1em}
\caption{\scriptsize{\textbf{Instruction Page 4} }}
\vspace{-2em}
\label{fig:interface5}
\end{center}
\end{figure}

\begin{figure}[hbt]
\begin{center}
\includegraphics[width=0.9\linewidth]{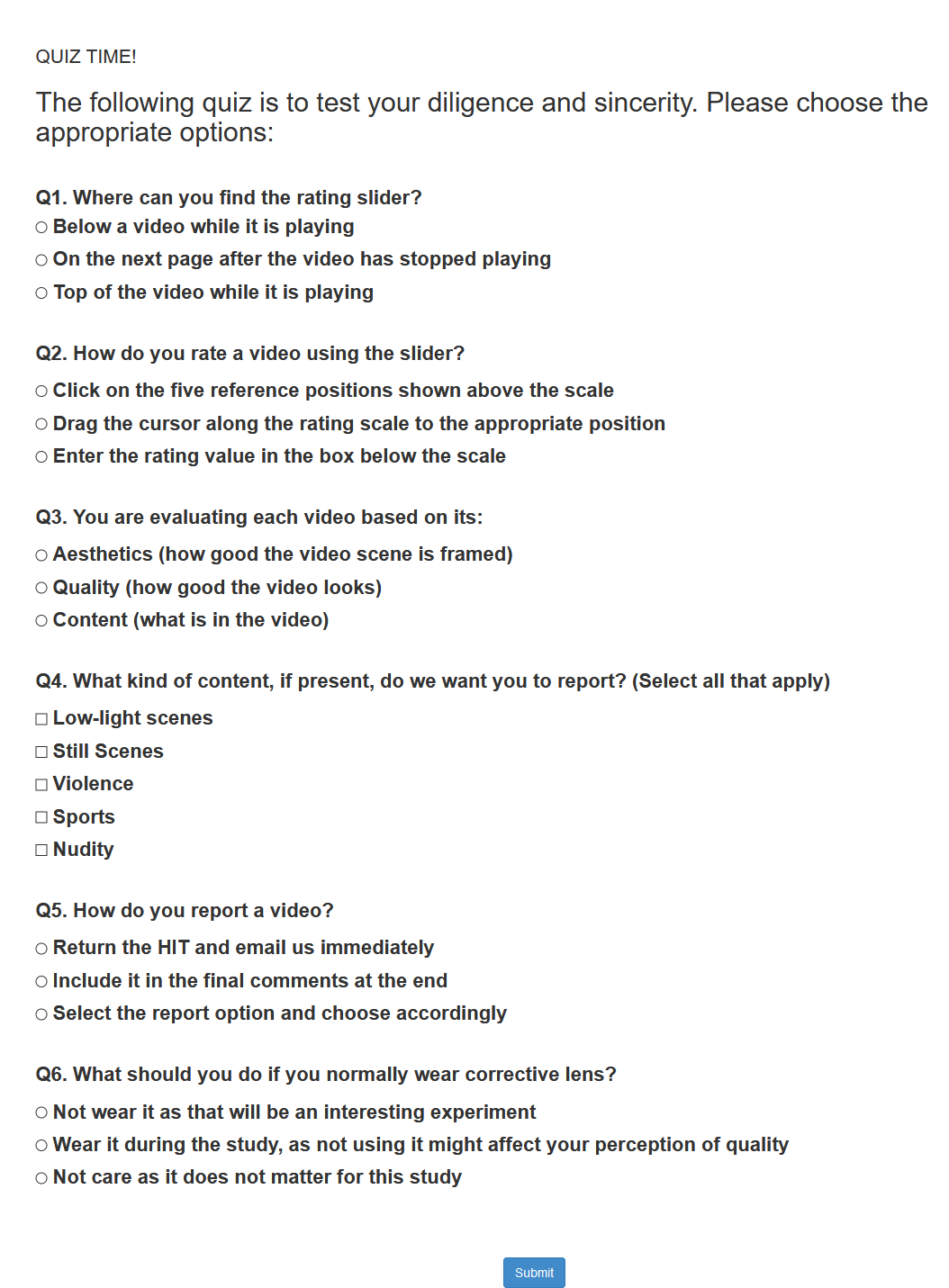}
\caption{\scriptsize{\textbf{Quiz} }}
\vspace{-2em}
\label{fig:interface_quiz}
\end{center}
\end{figure}

\begin{figure}[hbt]
\begin{center}
\includegraphics[width=1\linewidth]{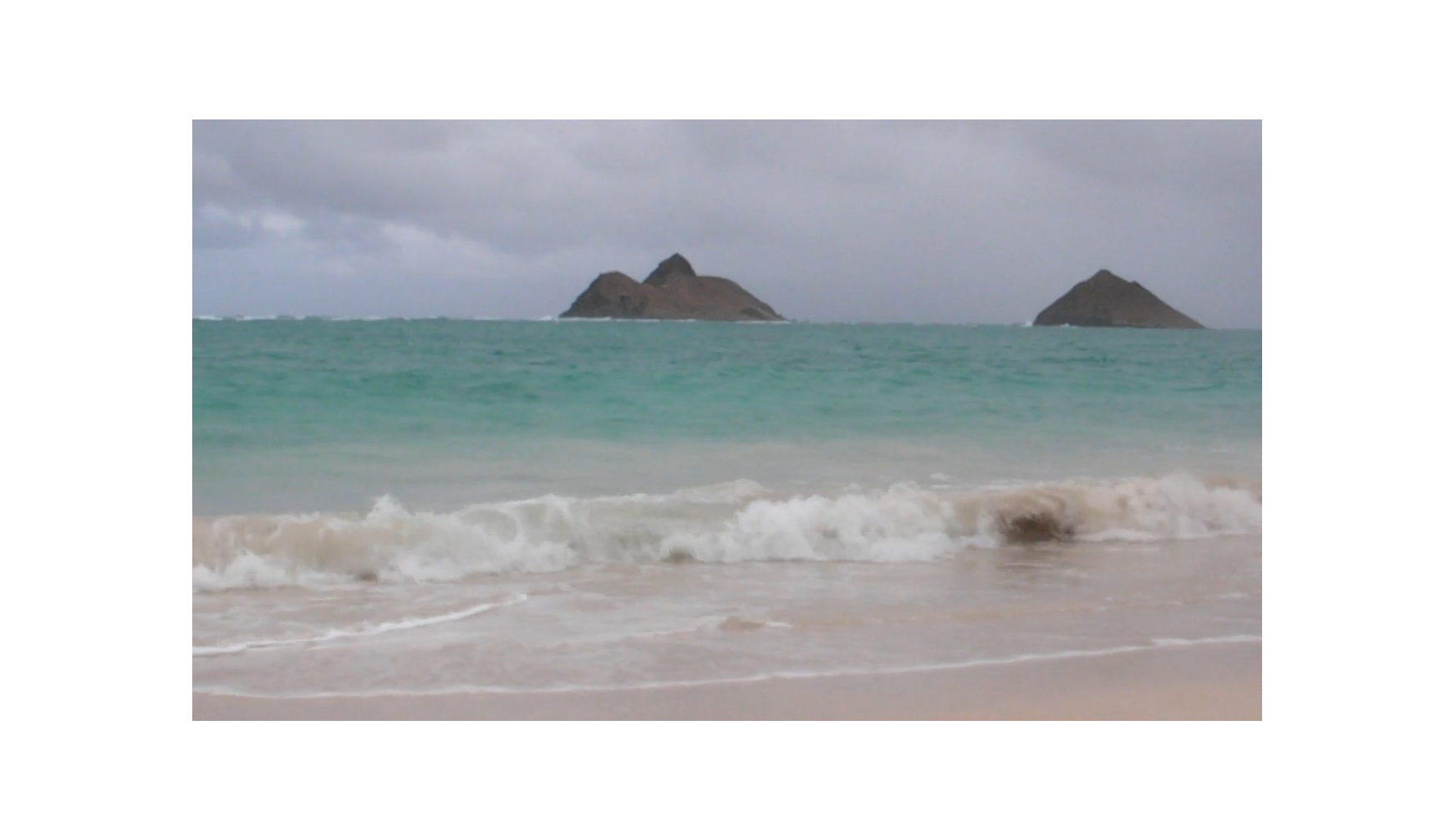}
\vspace{-1em}
\caption{\scriptsize{\textbf{Video Playback} }}
\vspace{-2em}
\label{fig:interface_vid}
\end{center}
\end{figure}

\begin{figure}[hbt]
\begin{center}
\includegraphics[width=1\linewidth]{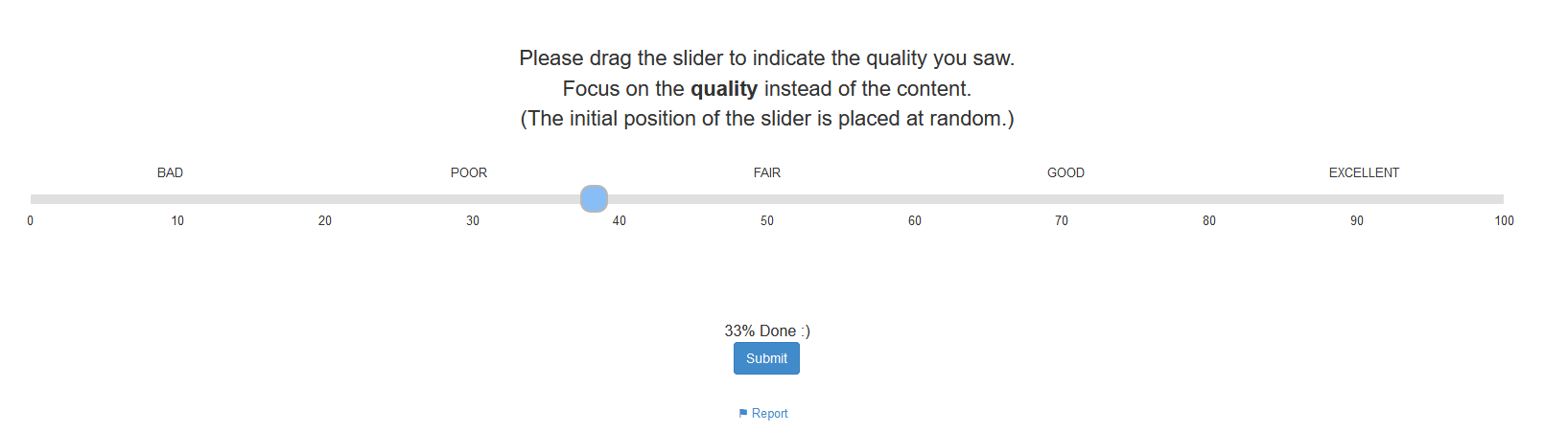}
\vspace{-1em}
\caption{\scriptsize{\textbf{Rating Slider} }}
\vspace{-2em}
\label{fig:interface_slider}
\end{center}
\end{figure}

\endgroup

\clearpage
\newpage



\end{document}